\documentclass[]{fairmeta}
\usepackage{xspace}
\usepackage{graphicx}
\usepackage{makecell}
\usepackage{multirow}
\usepackage[table]{xcolor}
\usepackage{caption}
\usepackage{subcaption}
\usepackage{amsfonts}
\usepackage{wrapfig}
\usepackage[most]{tcolorbox}
\usepackage{listings}  
\usepackage{sidecap, caption}

\definecolor{myboxcolor}{RGB}{245,245,245} 
\definecolor{myframe}{RGB}{0,0,128} 

\newtcolorbox{mybanner}{
  colback=myboxcolor,
  colframe=myframe,
  boxrule=1pt, 
  left=1pt,
  right=1pt,
  top=1pt,
  bottom=1pt,
}

\newtcolorbox{mybody}{
  colback=myboxcolor,
  colframe=myframe,
  boxrule=1pt, 
  left=1pt,
  right=1pt,
  top=1pt,
  bottom=1pt,
}

\definecolor{my_green}{RGB}{51,102,0}

\newcommand{\name}[1]{\textsc{HoneyBee}\xspace}

\newcommand{\virl}[1]{\texttt{ViRL}\xspace}
\newcommand{\mathllava}[1]{\texttt{Math-LLaVA}\xspace}
\newcommand{\ronevision}[1]{\texttt{R1-OneVision}\xspace}
\newcommand{\thinklite}[1]{\texttt{ThinkLite-VL-Hard}\xspace}
\newcommand{\vwi}[1]{\texttt{VisualWebInstruct}\xspace}
\newcommand{\llavacot}[1]{\texttt{LLaVA-CoT}\xspace}
\newcommand{\mmk}[1]{\texttt{MMK12}\xspace}
\newcommand{\ot}[1]{\texttt{OpenThoughts3}\xspace}

\newcommand{\mverse}[1]{\texttt{MathVerse}\xspace}
\newcommand{\mvista}[1]{\texttt{MathVista}\xspace}
\newcommand{\mvision}[1]{\texttt{MathVision}\xspace}
\newcommand{\mmmupro}[1]{\texttt{MMMU-Pro}\xspace}
\newcommand{\wemath}[1]{\texttt{We-Math}\xspace}
\newcommand{\dynamath}[1]{\texttt{DynaMath}\xspace}
\newcommand{\lvista}[1]{\texttt{LogicVista}\xspace}
\newcommand{\hbench}[1]{\texttt{HallusionBench}\xspace}
\newcommand{\tmath}[1]{\texttt{MATH500}\xspace}
\newcommand{\tgpqa}[1]{\texttt{GPQA}\xspace}

\newcommand{\qwen}[1]{Qwen2.5-VL\xspace}
\newcommand{\intern}[1]{InternVL-2.5\xspace}
\newcommand{\internthree}[1]{InternVL-3\xspace}

\newcommand{\lscout}[1]{\texttt{Llama4-Scout}\xspace}

\title{\name{}: Data Recipes for Vision-Language Reasoners}

\author[1,2,*]{Hritik Bansal}\author[1]{Devendra Singh Sachan}\author[2]{Kai-Wei Chang}\author[2]{Aditya Grover}\author[1]{Gargi Ghosh}\author[1]{Wen-tau Yih}\author[1]{Ramakanth Pasunuru}
\affiliation[1]{FAIR at Meta}
\affiliation[2]{University of California Los Angeles}

\contribution[*]{Work done at Meta}

\abstract{Recent advances in vision-language models (VLMs) have made them highly effective at reasoning tasks. However, the principles underlying the construction of performant VL reasoning training datasets remain poorly understood. In this work, we introduce several data curation approaches and study their impacts on VL reasoning capabilities by carefully controlling training and evaluation setups. We analyze the effects of context (image and question pair) sources, implement targeted data interventions, and explore scaling up images, questions, and chain-of-thought (CoT) solutions. Our findings reveal that (a) context source strategies significantly affect VLM performance, (b) interventions such as auxiliary signals from image captions and the inclusion of text-only reasoning yield substantial gains, and (c) scaling all data dimensions (e.g., unique questions per image and unique CoTs per image-question pair) consistently improves reasoning capability. Motivated by these insights, we introduce \name{}, a large-scale, high-quality CoT reasoning dataset with $2.5$M examples consisting $350$K image-question pairs. VLMs trained with \name{} outperform state-of-the-art models across model sizes. For instance, a \name{}-trained VLM with $3$B parameters outperforms the SOTA model and the base model by $7.8\%$ and $24.8\%$, respectively, on MathVerse. Furthermore, we propose a test-time scaling strategy that reduces decoding cost by $73\%$ without sacrificing accuracy. Overall, this work presents improved strategies for VL reasoning dataset curation research.}

\date{\today}
\correspondence{\email{hbansal@g.ucla.edu}}

\metadata[Code]{\url{https://github.com/facebookresearch/HoneyBee_VLM}}

\begin{document}

\maketitle

\begin{figure}[h]
\centering
    \begin{subfigure}[h]{0.5\textwidth}
        \centering
        \includegraphics[width=0.85\textwidth]{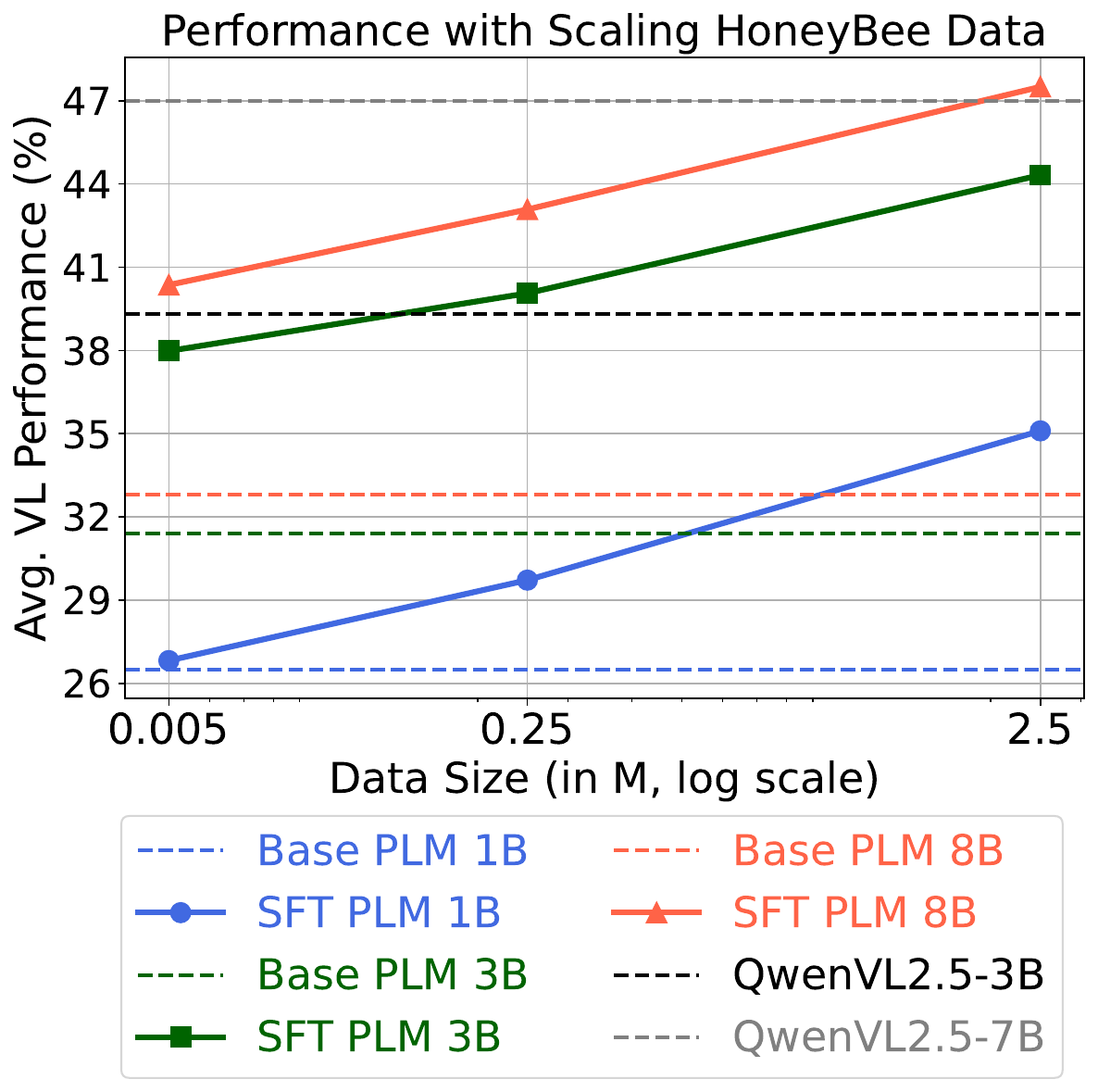}
        \caption{}
        \label{fig:summary_results}
    \end{subfigure}
    \begin{subfigure}[h]{0.48\textwidth}
    \centering
        \includegraphics[width=0.85\textwidth]{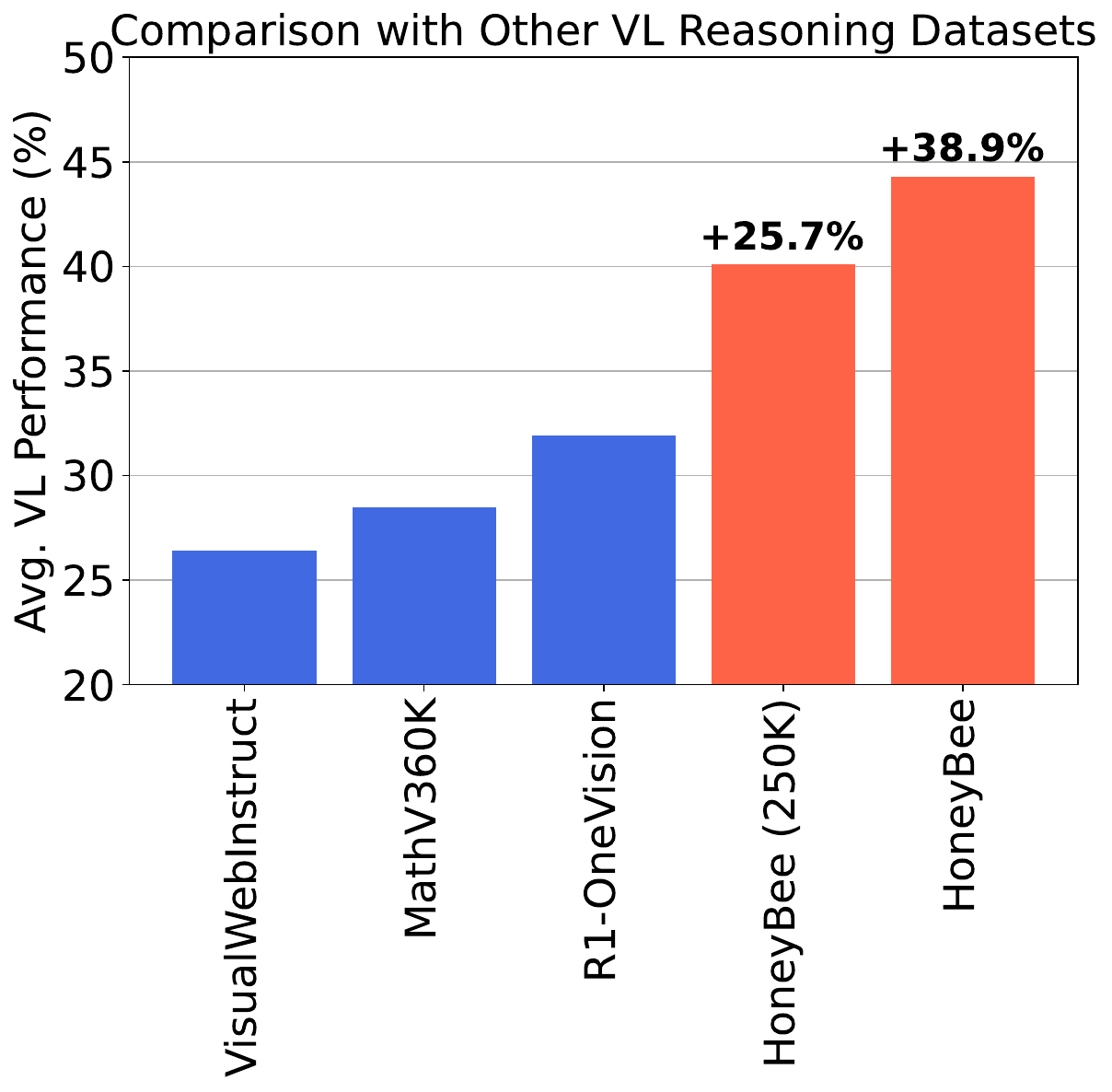}
        \caption{}
    \label{fig:comparison_with_existing}
    \end{subfigure}
    \caption{\small{\textbf{Summary of the results.} (a) We show that training with increasing amounts of our curated \name{} data leads to consistent accuracy improvements averaged across five VL reasoning tasks (MathVista, MathVerse, MathVision, MMMU-Pro, and We-Math) for several model sizes ($1$B to $8$B). (b) We report the relative gains achieved by training \name{} with PLM-3B compared to existing VL reasoning CoT datasets containing 250K–300K VL instances.}}
    \label{fig:pull_figure}
\end{figure}

\section{Introduction}
\label{sec:introduction}

Solving reasoning problems, such as those involving mathematics in visual contexts, is a crucial capability for AI models, powering many real-world applications such as visual data analysis \cite{masry2022chartqa,lu2023mathvista}, education \cite{baral2025drawedumath}, and scientific discovery \cite{trinh2024solving}. Recent vision-language models (VLMs), such as GPT-4o \cite{hurst2024gpt}, o3 \cite{OpenAI-o3}, Gemini-2.5 \cite{comanici2025gemini}, Llama-4 \cite{meta2025llama} achieve strong VL reasoning performance by training their pretrained models on high-quality synthetic chain-of-thought (CoT) data \cite{zelikman2022star}. In particular, the ability to generate CoTs (e.g., step-by-step solution) during problem-solving enable VLMs to utilize additional inference-time computation before providing the final answer \cite{wei2022chain,kojima2022large,guo2025deepseek}. Yet, the multimodal CoT datasets and their recipes used for training state-of-the-art VLMs are often proprietary, leaving several open questions about their design space.

\begin{table}[t]
\centering
\resizebox{\textwidth}{!}{%
\begin{tabular}{lccccccccccc}
\toprule
& \rotatebox{90}{\textbf{Average}} & \rotatebox{90}{\makecell{\textbf{MathVerse}\\(testmini,\\vision-only)}} & \rotatebox{90}{\makecell{\textbf{MathVista}\\(testmini)}} & \rotatebox{90}{\makecell{\textbf{MathVision}\\(testmini)}} & \rotatebox{90}{\makecell{\textbf{MMMU-Pro}\\(vision)}} & \rotatebox{90}{\makecell{\textbf{We-Math}\\(testmini)}} & \rotatebox{90}{\textbf{DynaMath}*} & \rotatebox{90}{\textbf{LogicVista}*}& \rotatebox{90}{\makecell{\textbf{Hall. Bench}*\\(image)}} & \rotatebox{90}{\textbf{MATH500}*\textsuperscript{\ddag}} & \rotatebox{90}{\makecell{\textbf{GPQA}*\textsuperscript{\ddag}\\(diamond)}} \\
\toprule
\textbf{1B model scale}&&&&&&&&&&&\\\hline
PLM-1B \cite{cho2025perceptionlm}& 25.9 & 17.8& 48.6 & 15.1& 15.8& 35.5 & 30.3& 23.0& 50.3& 15.2 & 7.1\\
\intern{}-1B \cite{chen2024expanding}& 27.6 & 18.7& 44.2 & 16.4& 16.2& 38.7 & 29.3& 20.5& 51.6& 27.8 & 12.1\\
\internthree{}-1B-Instruct \cite{zhu2025internvl3}& 28.3& 18.0& 35.0 & 13.2& 16.2& 37.5 & 28.5& 21.9& 56.0& 37.8 & 19.2 \\
\rowcolor{blue!20}PLM-HoneyBee-1B & \textbf{36.2} & \textbf{29.4}&\textbf{53.7}&\textbf{23.0}&\textbf{18.8}&\textbf{50.6}& \textbf{39.3} & \textbf{28.6} & \textbf{56.1} & \textbf{36.8}&\textbf{25.8}\\\hline
\textbf{3B-4B model scale} &&&&&&&&&&&\\\hline
PLM-3B \cite{cho2025perceptionlm}& 33.8& 18.0& 57.2 & 16.1& 19.5& 46.1 & 37.0& 33.5& 61.1& 30.4 & 18.7 \\
\intern{}-4B \cite{chen2024expanding}& 41.5 & 28.7& \textbf{61.8} & 24.7& \textbf{31.1}& 55.6 & 40.7& 32.1& 65.5& 49.4 & 25.3 \\
\qwen{}-3B-Instruct \cite{bai2025qwen25}& 42.6& 35.0& 58.9 & 23.7& 29.8& 49.2 & 42.5& \textbf{36.6}& \textbf{66.0}& 62.0 & 22.7 \\
\rowcolor{blue!20} PLM-HoneyBee-3B & \textbf{46.2} & \textbf{42.8}&61.2&\textbf{29.9}&28.4&\textbf{59.3} & \textbf{51.9} & \textbf{36.6} & 65.0  &59.4 & \textbf{27.7}\\
\textbf{7B-8B model scale}&&& &&&& & & & &\\\hline
PLM-8B \cite{cho2025perceptionlm}& 34.6& 19.3& 59.3 & 17.1& 20.5& 47.9 & 36.7& 30.8& 64.0& 34.0 & 16.2 \\
\intern{}-8B \cite{chen2024expanding}& 41.4 & 27.3& 61.5 & 21.4& 32.0& 56.6 & 41.9& 26.6& 63.8& 57.0 & 26.3\\
\internthree{}-8B-Instruct \cite{zhu2025internvl3}& 45.1 & 35.3& 61.8 & 19.4& 35.8& 55.7 & 51.2& 36.2& 65.5& \textbf{69.6} & 20.2 \\		
\qwen{}-7B-Instruct \cite{bai2025qwen25}& 48.5& 42.0& 67.5 & \textbf{27.6}& \textbf{37.1}& 61.1 & 51.3& 39.9& 67.4& 64.8 & 26.3 \\		
\rowcolor{blue!20}PLM-HoneyBee-8B & \textbf{49.8}& \textbf{43.0}&	\textbf{68.2}&	26.3	&33.8	&\textbf{66.1}& \textbf{53.3}&	\textbf{41.3}&	\textbf{68.8} &63.6	&\textbf{33.3}  \\
\bottomrule
\end{tabular}%
}
\caption{\small{\textbf{Performance of VL reasoners trained with \name{} data.} We compare the accuracy of PLMs trained with the \name{} data on diverse downstream evaluation datasets. We find that models trained on \name{} achieve best-in-class performance across model sizes. Task-specific subsets or splits are indicated in brackets `()'. Datasets that were unseen during the data curation process are marked with *, and text-only reasoning datasets are marked with \textsuperscript{\ddag}.}}
\label{tab:main_table}
\end{table}

Several prior works have shown that supervised finetuning (SFT) with the quality of CoT data is crucial for LLM (text-only) reasoning performance \cite{guha2025openthoughts, bercovich2025llamanemotronefficientreasoningmodels, muennighoff2025s1, sky_t1_2025}, and also serves as a key foundation for subsequent RL training \cite{liu2025acereason}. However, there is a major gap in our understanding of how high-quality CoT datasets are constructed for VL reasoning, where a model needs to integrate information from several modalities (visual content in images and text content in questions) to provide accurate answers. Specifically, prior work on VL reasoning does not explore the breadth of design choices and suffers from several challenges. Firstly, the impact of context (image and question pairs) from diverse data sources remains unclear. For instance, \mathllava{}~\cite{shi2024math} and \llavacot{}~\cite{xu2024llava} curate different data distributions from existing image QA datasets and employ their own custom CoT generation strategies. Thus, it is uncertain how much of the reasoning performance of models trained on these datasets can be attributed to the quality of the context.
Secondly, prior work~\cite{li2025vision,he2025deepmath,chen2025g1} suggests that targeted data interventions—such as visual perturbations and difficulty filtering—can enhance model perception and problem-solving. However, there has been limited exploration of other interventions that could further improve data quality.
Thirdly, while scaling up the training data is known to enhance reasoning performance~\cite{guha2025openthoughts}, it remains unclear whether VL reasoning data should be scaled along specific axes or across all axes, such as the number of images, the number of questions per image, and the number of CoTs per (image, question) pair.

\begin{figure}[t]
    \centering
    \includegraphics[width=\linewidth]{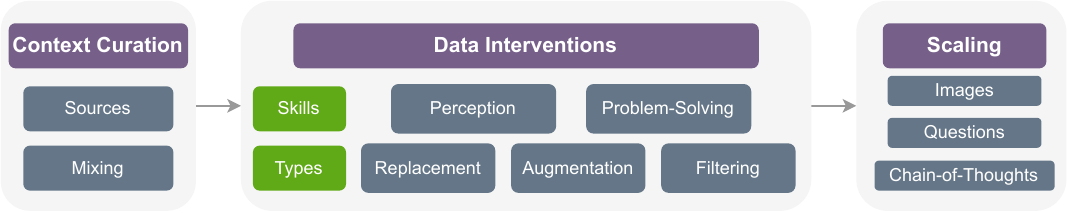}
    \caption{\small{\textbf{Overview of the data curation pipeline.} In this work, we curate the context (image, question) from diverse sources and assess the impact of their mixing. Further, we curate a set of data interventions that target diverse skills and types. Subsequently, we study the impact of scaling along different data axes. These insights lead to the creation of our large-scale \name{} dataset.}}
    \label{fig:pipeline}
\end{figure}

Importantly, there is a lack of fair and robust comparisons between diverse design decisions. For fairness, the direct effect of a design choice should be measured by fixing the training setups (e.g., SFT starting from identical models) and the evaluation protocol. For robustness, models should be trained at multiple scales (such as 3B and 8B) and evaluated on a collection of datasets rather than a single dataset.
To address these critical questions in VL reasoning data design, we adopt a comprehensive and scalable approach to identify the key factors in dataset curation (Figure~\ref{fig:pipeline}). This effort culminates in the creation of \textbf{\name{}}, a high-quality and one of the largest VL reasoning CoT datasets, comprising \textbf{2.5M} (image, question, CoT) tuples.

In our data curation process, we first study the impact of diverse contexts (i.e., image and question pairs) acquired from several VL reasoning datasets and rank them based on the performance of multiple VLMs (i.e., 3B and 8B models) on a battery of evaluation datasets. Interestingly, we observe that the choice of source datasets can lead to significant differences in model performance, with up to a 4\% difference in average accuracy across evaluation datasets (\S\ref{sec:context_curation}). In the next stage, we create a list of data enhancement strategies, including \textit{visual perturbation}, \textit{text-rich images}, \textit{perceptual redundancy}, \textit{shallow perception}, \textit{caption-and-solve}, \textit{text-only reasoning}, \textit{increased distractors}, \textit{length} and \textit{difficulty} filtering, applied on top of the best-performing dataset from the previous stage (Figure~\ref{fig:interventions}). Our experiments show that several data interventions fail to outperform the baseline dataset, highlighting their limited practical value despite strong motivations. Importantly, we reveal that auxiliary signals from image captions (\textit{caption-and-solve}) and augmenting the VL reasoning data with text-only reasoning data lead to major improvements across diverse benchmarks (\S\ref{sec:data_interventions}). In our experiments, we find that model performance improves with scaling all dimensions (images, questions, and CoTs) in the reasoning data (\S \ref{sec:scaling_million}). 

Motivated by these findings, we construct the \name{} dataset, and train VLMs of several sizes ($1$B–$8$B) on the \name{} dataset. We show that reasoning performance strongly scales with the amount of training data and outperforms existing state-of-the-art instruction-tuned VLMs, indicating at the quality and scalability of our dataset (\S \ref{sec:training_with_honeybee}). In particular, PLM-\name{}-1B achieves a relative performance improvement of $28$ percentage points (pp) over \internthree{}-1B-Instruct, averaged across ten evaluation datasets (Table~\ref{tab:main_table}). Furthermore, PLM-\name{}-3B and PLM-\name{}-8B achieve relative gains of $8.4$pp and $2.7$pp over \qwen{}-3B-Instruct and \qwen{}-8B-Instruct, respectively.
Ultimately, we also propose an efficient decoding strategy for test-time scaling that enables the generation of multiple solutions from the \name{}-trained VL reasoners, using $73\%$ fewer inference tokens without any loss in performance (\S~\ref{sec:efficiency_tts}). Our thorough experiments yield several insightful findings that lay the foundation for curating the next generation of VL reasoning datasets.
\section{Preliminaries}
\label{sec:preliminaries}

In this work, we focus on the curation of high-quality, large-scale synthetic data to enable strong vision-language reasoning capabilities. Let $\mathcal{D} = \{(I_j, Q_j, A_j)\}_{j=1}^{N}$ denote the source vision-language reasoning dataset of size $N$, where each entry consists of an image $I_j$, a corresponding question $Q_j$, and an optional final answer $A_j$.\footnote{When there is no final answer, we can set $A_j = \phi$.} Further, let $\mathcal{G}$ be a synthetic data generator that outputs a textual chain-of-thought (CoT) to solve the questions about the images in $\mathcal{D}$, such that $C_j = \mathcal{G}(I_j, Q_j)$. In particular, the CoT $C_j$ is composed of several reasoning steps, including step-by-step solutions, planning, self-verification, and self-reflection behaviors \citep{team2025kimi}, denoted as $S_j$. This is followed by a predicted final answer $P_j$ to the given question about the image. Thus, $C_j = [S_j; P_j]$, where $;$ denotes concatenation in the raw text space. Thus, We represent the synthetic data as $\mathcal{D}_{\mathcal{G}} = \{(I_j, Q_j, C_j, A_j)\}_{j=1}^{N}$.

Given synthetic data, we train a VLM ($p_\theta$), using a supervised finetuning objective: $\mathbb{E}_{(I_j, Q_j, C_j) \in \mathcal{D}_\mathcal{G}}[\log p_\theta(C_j \mid I_j, Q_j)]$, i.e., maximizing the probability of the problem-solving CoT given the image and question as context. Post-training, the VLM will perform step-by-step reasoning before generating its predicted answer, thus utilizing additional test-time compute \citep{wei2022chain}. Overall, the goal of high-quality synthetic data is to train performant VL reasoners that can solve novel problems from diverse downstream tasks such as geometry, function plots, and charts \citep{lu2023mathvista}. In this work, we focus on the paradigm of training on smaller VLMs on synthetic data from larger generator models. This approach is more compute-efficient, more popular \cite{shi2024math,guha2025openthoughts,bercovich2025llamanemotronefficientreasoningmodels,li2024numinamath}, and allows for comprehensive training runs on diverse synthetic data distributions. In practice, $p_\theta$ can be a weaker VLM reasoner than the generator \citep{shi2024math}, a setup commonly referred to as knowledge distillation \citep{hinton2015distilling}, where a strong teacher guides a weak student. Alternatively, the generator itself can be used as the student, a process known as self-improvement \citep{singh2023beyond}. Prior research has also explored training a stronger reasoner with data from a weaker one, referred to as weak-to-strong reasoning \citep{burns2023weak,yang2024weak,bansal2024smaller}.

\section{Data Curation Pipeline}
\label{sec:data_curation}

We outline our vision-language reasoning data curation pipeline (Figure \ref{fig:pipeline}), which consists of multiple stages: (a) \texttt{context curation}, which assesses the impact of diverse context (image, question) data sources, as well as their mixing (\S\ref{sec:context_curation}); (b) \texttt{data interventions}, which aim to enhance perception and problem-solving skills to enable strong VL reasoning capabilities (\S\ref{sec:data_intervention_explain}); and (c) \texttt{scaling}, which studies the impact of scaling diverse components of the reasoning data (\S\ref{sec:scaling_diverse_modalities}).

\subsection{Context Curation}
\label{sec:context_curation}

The quality of context i.e., image, question pair, is crucial for determining the knowledge, and reasoning skills imparted to the VL reasoner. For instance, a VL reasoner exposed to diverse geometric images and questions will excel in downstream geometry problem-solving \citep{zhang2024mavis}. Thus, we aim to curate contexts that allow the resulting VL reasoner to achieve good performance on several downstream reasoning tasks.

\paragraph{\textbf{Sourcing.}}
In the past few years, several CoT datasets have been proposed to enable strong vision-language (VL) reasoning capabilities \citep{shi2024math, jia2025visualwebinstruct, xu2024llava}. However, the reported improvements are a combination of several factors, such as the contexts present in the dataset, custom training algorithms, and the quality of the generator model. Furthermore, the contexts in different datasets originate in diverse ways. In particular, the contexts in reasoning datasets like \mathllava{} and \ronevision{} \citep{shi2024math, yang2025r1} are composed of several existing expert-curated and task-specific datasets, such as Geometry-3K \cite{lu2021inter}, IconQA \cite{lu2021iconqa}, CLEVR-Math \cite{lindstrom2022clevr}, while \virl{} \citep{vl-rethinker} integrates several VL datasets \cite{du2025virgo} and performs a customized cleaning process. Similar to prior data curation work \citep{guha2025openthoughts, li2024datacomplm}, we fix the training algorithm (e.g., supervised finetuning) and the CoT generator model to analyze the \textit{direct effect} of the contexts in individual datasets on training performant VL reasoners. 

First, we list several widely adopted VL reasoning datasets consisting of (image, question, final answer) tuples. 
Next, we remove instances from these datasets that contain images with identical perceptual hashes\footnote{\url{https://github.com/idealo/imagededup}} to any of the images in the evaluation VL reasoning tasks~\citep{zhu2023multimodal}. Second, we prompt our generator model to produce a CoT to solve the (image, question) pairs in these datasets. Importantly, we consider small, similarly sized subsets of these datasets to control for the impact of data quality, rather than data quantity, on model performance. Similar to \citep{guha2025openthoughts}, this approach also facilitates faster training iterations. Further, we also create a filtered version of each dataset by discarding the CoTs that do not lead to the correct final answer. This step ensures that we do not penalize a dataset when our generator produces an incorrect solution. Third, we train multiple VLMs (e.g., 3B and 8B models) on different context sources (and their filtered versions) and evaluate them on several downstream VL reasoning tasks. Finally, we rank each individual source dataset based on the average performance across multiple downstream tasks and model trainings.

\begin{figure}[t]
    \centering
    \includegraphics[width=\linewidth]{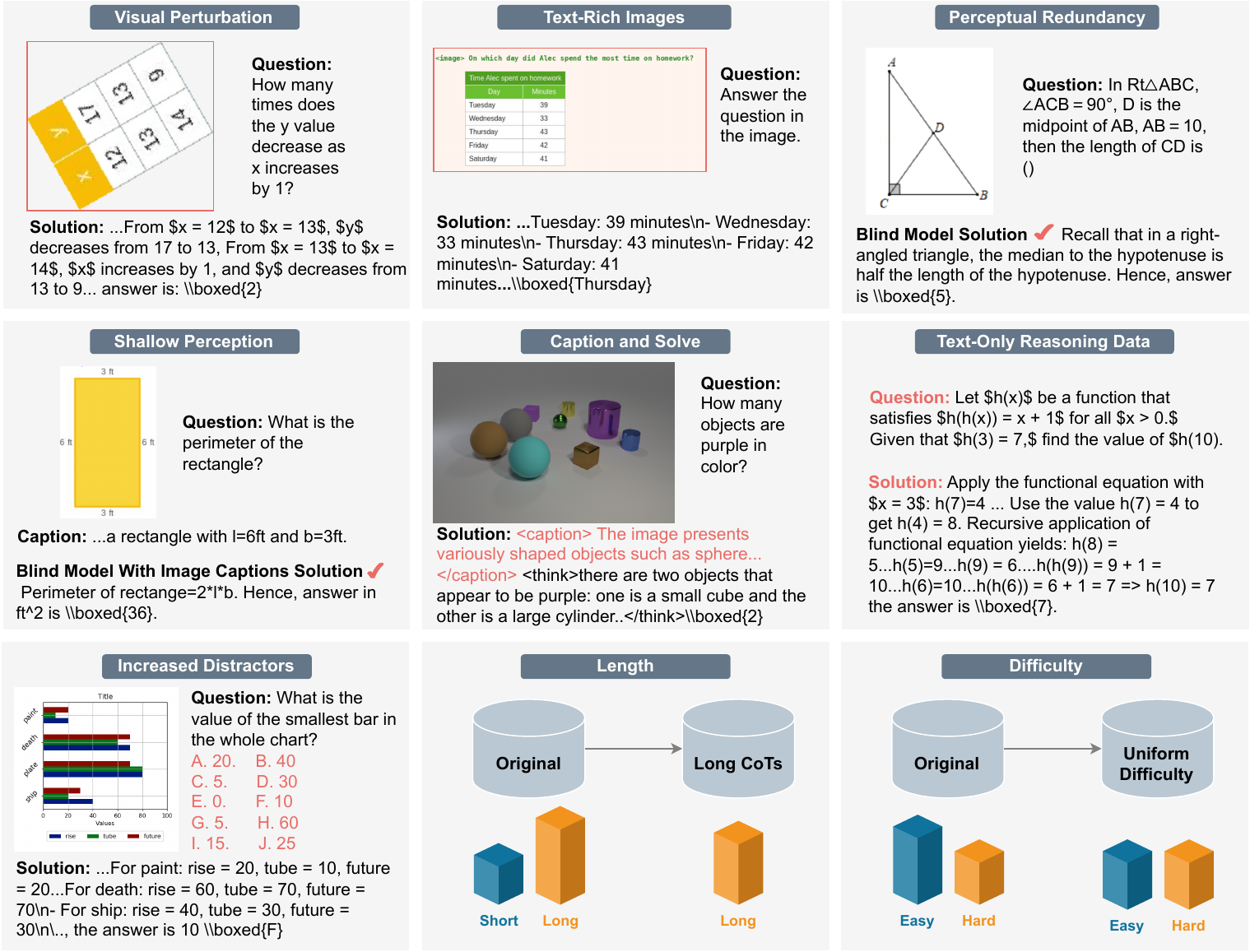}
    \caption{\small{\textbf{Overview of data intervention strategies.} In this work, we curate a diverse set of data interventions to enhance the quality of VL reasoning CoT data. These methods target a range of skills, such as perception and problem-solving capabilities. We present details for each intervention in \S \ref{sec:data_intervention_explain}.}}
    \label{fig:interventions}
\end{figure}

\paragraph{\textbf{Mixing.}}
Here, we consider mixing as an additional strategy to combine the strengths of the top-performing data sources. Specifically, prior work in LLM reasoning \cite{guha2025openthoughts, lambert2024tulu} has shown that mixing can yield superior datasets compared to using individual sources alone. To this end, we mix the data from the top-2, top-4, and all datasets from the previous stage. Subsequently, we take an equally sized subset from each mixture to control for dataset size and focus solely on mixture quality. We then train multiple VLMs on these mixtures and evaluate them on a battery of VL reasoning tasks. Finally, we assess whether any of the mixtures achieve better average performance than the best-performing individual dataset.

\subsection{Data Interventions}
\label{sec:data_intervention_explain}

Starting with the best data mixture from the previous stage, we assess whether its quality can be further enhanced through targeted data interventions (Figure \ref{fig:interventions}). Specifically, these interventions aim to improve particular skills of the VL reasoner, including \textit{perception} and \textit{problem-solving} capabilities. Enhanced perception is crucial for robust visual understanding of image content within the model's context, and strong problem-solving ability is essential for accurate step-by-step solutions and calculations. 

In addition, the data intervention strategies can affect the original VL CoT reasoning data in various ways, including \textit{replacement}, \textit{augmentation}, and \textit{filtering}. The replacement strategy substitutes part (or all) of the original dataset with higher-quality variants, without changing the overall size of the dataset \cite{maini2024rephrasing}. In contrast, the augmentation strategy increases the amount of training data by either adding transformed variants of the original data \cite{zhang2017mixup} or introducing external knowledge \cite{bansal2023leaving}. Finally, we also consider diverse filtering scenarios, which improve the quality of the original data by removing lower-quality instances according to predefined rules or classifiers \cite{gadre2023datacomp}. 

\subsubsection{Perception Enhancement}
\label{sec:percept_data_interventions}

First, we will outline the various data interventions for enhancing the \textit{perception} of the VL reasoner:

\paragraph{\textbf{Visual Perturbation.}} This intervention creates a perturbed version of the original image in the reasoning dataset to enhance the perceptual robustness of the VLM \citep{li2025vision}. Starting from an instance in the  synthetic reasoning dataset, $\{I_j, Q_j, C_j\} \in \mathcal{D}_\mathcal{G}$, we apply one of three transformations, $\mathcal{T} =$ \{\textit{rotation}, \textit{distractor concatenation}, \textit{dominance-preserving mixup}\}, randomly to the image in each instance such that $I'_j = h(I_j)$, where $h$ is a transformation from $\mathcal{T}$. We test the impact of this method in two ways: (a) as a \textit{replacement}, where we perturb the images in part of the dataset while leaving the other part untouched, and (b) as an \textit{augmentation}, where we mix the transformed dataset with the original dataset.

\paragraph{\textbf{Text-Rich Images.}} It is crucial for VLMs to comprehend textual information embedded in images to perform accurate reasoning \citep{yue2024mmmu,wadhawan2024contextual}. To enhance the model's ability to read and integrate textual information within images, we programmatically transform each (image, question) pair from the original data into a new image. Specifically, the new dataset $\mathcal{D}_{TR}$ contains only the transformed text-rich image and the CoT from the original dataset. In particular, we embed the question in diverse font styles and colors, overlaying it with the original image on a blank background with various colors. Formally, $I'_j = Render(I_j, Q_j)$, thus, the new dataset is $\mathcal{D}_{TR} = \{I_j', C_j\}_{j=1}^{N}$. Similar to visual perturbation, we assess the usefulness of this method as both a \textit{replacement} and an \textit{augmentation} strategy.

\paragraph{\textbf{Perceptual Redundancy.}} Prior work~\cite{zhang2024mathverse} has shown that VLMs often rely on the visual content embedded within textual questions for answer deduction, rather than truly understanding the context image. Additionally, we observe that some textual questions can be answered accurately without access to the image, using only the model’s existing knowledge. For instance, a strong VLM can utilize its pretrained knowledge to answer questions (e.g., \texttt{What will lead to an increase in the population of deer? i. increase in lion, ii. decrease in plants, \textbf{iii. decrease in lion}, iv. increase in pika}) without access to the food web image \cite{kembhavi2016diagram}. To encourage greater reliance on visual inputs, we \textit{filter} the original synthetic dataset by removing instances where the generator model leads to the correct final answer without access to the image; that is, we exclude examples where the predicted answer $\hat{P_j} = \mathcal{G}(Q_j)$ matches the ground-truth answer $A$ (i.e., $\hat{P_j} = A_j$).\footnote{In our experimental setup, we use the same foundation model as CoT generator as well as caption generator. Thus, we take the liberty to denote image captioner as $\mathcal{G}$.} 

\paragraph{\textbf{Shallow Perception.}} In the existing VL datasets, there are several instances that can be solved by querying the generator to solve them with access to the question and an image caption. Since image caption contains shallower information than all the visual content in the original image, such instances will not require deep visual insights in the reasoning process. Thus, we filter the original dataset where the generator model leads to the correct final answer with access to the question and image caption but not the image; that is, we exclude examples where the predicted answer $\hat{P_j} = \mathcal{G}(Q_j, I_{j}^{cap})$ matches the ground-truth answer $A_j$ (i.e., $\hat{P_j} = A_j$). In our experiments, the image caption  $I_j^{cap} = \mathcal{G}(I_j)$ is also generated by the same generator model.

\paragraph{\textbf{Caption and Solve.}} We explore an active approach to enhancing the visual understanding capabilities of the VL reasoner by providing auxiliary visual signals from the image caption $I_j^{cap}$ from the stronger generator model. Specifically, we augment the original dataset’s CoT by including the image caption in the training data, i.e., $C_j' = [I_j^{cap}; C_j]$, where $;$ denotes concatenation in the raw text space. The transformed reasoning dataset is then $\mathcal{D}_{CS} = \{I_j, Q_j, C'_j\}_{j=1}^{N}$. By design, this method also increases the length of the CoT, $|C'_j| > |C_j|$, which facilitates the use of inference-time compute. We provide more information about different configurations in Appendix \ref{app:data_interventions}.

\subsubsection{Problem-Solving Skill Enhancement}
\label{sec:prob_solve_data_interventions}

Next, we will outline the various data interventions for enhancing the \textit{problem-solving} skills of the VL reasoner:

\paragraph{\textbf{Text-Only Reasoning.}} Through this intervention, we aim to enhance problem-solving skills by exposing the VLM to novel problems and solutions from a high-performing text-only reasoning dataset \citep{guha2025openthoughts}. This strategy serves multiple purposes: (a) it allows the VLM to learn useful skills from existing unimodal reasoning data via cross-task transfer, and (b) it makes the VLM a more general-purpose reasoner, enabling it to solve textual reasoning problems too beyond VL reasoning problems. In practice, this method acts as an \textit{augmentation}, where the final dataset is a concatenation of the original VL reasoning data and the high-performing text-only reasoning data, i.e., $\mathcal{D}_\mathcal{G}^{VL+Text} = \mathcal{D}_\mathcal{G}^{VL} \cup \mathcal{D}_\mathcal{G}^{Text}$. For consistency, we re-annotate the problems from the text-only reasoning data to obtain unimodal CoTs from the same generator model.

\paragraph{\textbf{Increased Distractors.}} Prior work \citep{wang2024mmlu,kazemi2025big,yue2024mmmu} has shown that introducing distractors in reasoning tasks, specifically by increasing the number of options (e.g., from four to ten), makes them harder for reasoning models to solve. Motivated by this observation, we explore a strategy to transform the original questions to have ten options and modify the original CoT accordingly; that is, $(Q_j', C_j') = \text{LLM}(Q_j, C_j)$. This method aims to encourage the VLM not to rely on chance when answering complex problems and to boost model robustness. We test the impact of this method in two ways: (a) as a \textit{replacement}, where part of the original dataset is transformed with distractors while the other part remains untouched, and (b) as an \textit{augmentation}, where we mix the transformed dataset with the original dataset.

\paragraph{\textbf{Length.}} In this method, we aim to encourage a reasoner to generate longer chains of thought (CoTs). This approach allows the model to utilize additional test-time compute for more detailed visual comprehension, stepwise solutions, planning, and re-evaluation during problem-solving. To achieve this, we compute the distribution of CoT lengths in the original reasoning dataset and split it into two equal halves: those with less than and those with more than the median CoT length. Subsequently, we train the models on the longer half to promote extended reasoning behavior in downstream tasks.

\paragraph{\textbf{Difficulty.}} Reasoning datasets are typically skewed toward easier problems, as these are easier to acquire at scale compared to more difficult problems, which only a limited number of experts can solve \citep{he2025deepmath}. To increase the difficulty of our dataset, we first classify all instances into diverse difficulty levels ($1$: beginner, $2$–$3$: AMC, $4$: intermediate level AIME, $>4$: Olympiad level) using an LLM and a predefined rubric.\footnote{\url{https://artofproblemsolving.com/wiki/index.php/AoPS_Wiki:Competition_ratings}} We then filter the dataset to ensure that the representation for each level is roughly balanced. This enables the model to learn from harder examples and tackle more complex VL reasoning problems in downstream tasks.

We present additional details about the data intervention strategies in Appendix \ref{app:data_interventions}. In theory, all the data intervention strategies listed above should enhance VL reasoning behaviors. However, an intervention strategy may steer the model toward behaviors that benefit a specific evaluation dataset at a particular model scale. Thus, outperforming a simple baseline of training on the original reasoning dataset across various downstream VL reasoning tasks and multiple model scales remains a non-trivial challenge. Therefore, it is essential to compare these strategies under identical conditions (e.g., training and evaluation) to identify those that consistently improve performance across the board.

\subsection{Scaling Diverse Data Axes}
\label{sec:scaling_diverse_modalities}

Scaling the amount of reasoning CoT data has consistently helped in enhancing the LLM performance on several downstream tasks \citep{guha2025openthoughts,toshniwal2024openmathinstruct2}. However, there has been limited exploration on the impact of scaling data for VL reasoning. In addition, the introduction of a new modality (image) in VL space adds a new dimension to scaling training data. Hence, we study the impact of scaling diverse axes (e.g., image, question, and CoT) on the reasoning capabilities of the VLM. Here, we explain the method to study the scaling behaviors starting from a dataset containing $N$ unique images, and corresponding question and CoT $\mathcal{D}_{N} = \{I_j, Q_j, C_j\}_{j=1}^{N}$. We drop the subscript $\mathcal{G}$ for simplicity.

\paragraph{\textbf{Scaling Images.}} We create three additional subsets of the original dataset, each of increasing size, resulting in four datasets in total: $[\mathcal{D}_{N/8}, \mathcal{D}_{N/4}, \mathcal{D}_{N/2}, \mathcal{D}_N]$. Here, $\mathcal{D}_{N/8}$ indicates that we randomly select $N/8$ instances from the original dataset. As we increase the data size, we are effectively scaling the number of unique images in the dataset. Subsequently, we train several VLMs on these datasets and study whether the average performance on downstream tasks improves with increased data scale.

\paragraph{\textbf{Scaling Questions.}} We study the impact of synthesizing novel questions by using the existing questions as seeds. In particular, we prompt the question generator to create new questions that are reasonable, solvable, and of similar difficulty to the original questions, conditioned on the image and question \citep{toshniwal2024openmathinstruct2}; i.e., we generate $Q'_{j,i} = \mathcal{G}(I_j, Q_j)$, where $Q'_{j,i}$ is the $i^{\text{th}}$ synthetic question for $(I_j, Q_j)$ using the question generator $\mathcal{G}$.\footnote{In our experiments, we use the same foundation model for both CoT generation and question generation; hence, we denote the question generator as $\mathcal{G}$.} Subsequently, we create a CoT for the newly synthesized questions using the generator model. We then merge this synthetic data with the original data, thereby scaling the dataset by having multiple questions for a given image. In our experiments, we construct four subsets of synthetic reasoning data: $[\mathcal{D}_{N/8}(n_q=1), \mathcal{D}_{N/8}(n_q=2), \mathcal{D}_{N/8}(n_q=4), \mathcal{D}_{N/8}(n_q=8)]$, where $\mathcal{D}_{N}(n_q=k) = \{I_j,Q_j,C_j\}_{j=1}^N \cup \{I_j, \{Q'_{j,i}, C_{j,i}\}_{i=1}^{i=k-1}\}_{j=1}^{j=N}$ and $C_{j,i} = \mathcal{G}(I_j,Q'_{j,i})$. If $n_q = 4$, then three questions are newly synthesized, and the remaining question comes from the original data. We present the prompt used to generate new questions and provide qualitative examples 
in Appendix \ref{app:prompts} and Appendix \ref{app:qualitative_examples}, respectively.

\paragraph{\textbf{Scaling CoTs.}} We study the impact of increasing the number of CoT traces for a given image and question pair in the reasoning dataset. This approach exposes the VL reasoner to diverse problem-solving strategies for a given problem. Specifically, we create four subsets of the dataset with an increasing number of CoTs per (image, question) pair: $[\mathcal{D}_{N/8}(n_c=1), \mathcal{D}_{N/8}(n_c=2), \mathcal{D}_{N/8}(n_c=4), \mathcal{D}_{N/8}(n_c=8)]$, where $\mathcal{D}_{N}(n_c=k) = \{I_j, Q_j, \{C_{j,i}\}_{i=1}^{k}\}_{j=1}^{N}$. Ultimately, we combine the findings from context curation, data interventions, and scaling diverse modalities experiments to create one of the largest VL reasoning datasets, \name{} (see \S\ref{sec:scaling_million} for more details).

\section{Experimental Setup}
\label{sec:setup}

\paragraph{\textbf{Training Data.}} Our goal is to collect a diverse datasets as sources of (image, question, final answer) tuples. This data will be used to create the multimodal CoT with our generator model. Specifically, we use six source datasets: \virl{} \cite{vl-rethinker}, \mathllava{} \cite{shi2024math}, \ronevision{} \cite{yang2025r1}, \thinklite{} \cite{wang2025sota}, \llavacot{} \cite{xu2024llava}, and \mmk{} \cite{meng2025mm}. We then perform decontamination to remove any identical images from the evaluation datasets using exact deduplication with the pHash algorithm. We cap the number of instances at $50$K to focus on the impact of data quality in our data curation experiments. Subsequently, we train VL reasoners for $5$ epochs across all experiments, and select the best-performing checkpoint (out of 5) based on the average performance across five downstream tasks. More details about the data statistics are provided in Appendix Table \ref{apptab:more_details_contexts}.

\paragraph{\textbf{Generator Model.}} We fix the CoT generator model to a performant VLM, \lscout{} \cite{meta2025llama}. Specifically, it is an open-weights model consisting of $109$B total parameters, of which $17$B are active. This model enables efficient inference on a single A100 node using vLLM \cite{vllmllama}. We use a non-API model to avoid changes in the quality over time \cite{chen2023chatgpt}, and powerful VLM that the community can host locally with accessible compute.

\paragraph{\textbf{Models.}}  We train the $3$B and $8$B Perception Language Models (PLMs) \cite{cho2025perceptionlm} for all data curation experiments. PLMs cannot generate vision-language CoTs, due to lack of appropriate instruction data. Thus, they serve as good base models that can be converted into VL reasoners with high-quality reasoning CoT data. Once our final data is ready, we also train much smaller VL reasoners (i.e., PLM-1B). We present more details about the training setup in Appendix \ref{app:training_setup}. 


\paragraph{\textbf{Evaluation.}} For our data curation experiments, we choose \textit{five} VL reasoning downstream tasks as validation datasets for hill-climbing. These include \mverse{} (testmini, vision-only subset) \cite{zhang2024mathverse}, containing $788$ examples on geometry and functions; \mvista{} (testmini) \cite{lu2023mathvista}, containing $1000$ examples on diverse visual scenarios (e.g., tables, functions, geometry, papers, and IQ tests); \mvision{} (testmini) \cite{wang2024measuring}, containing $304$ examples sourced from math competitions of various difficulties; \mmmupro{} (vision) \cite{yue2024mmmu}, containing $1730$ examples from diverse subject areas (e.g., art and design, science, humanities, and engineering); and \wemath{} (testmini) \cite{qiao2024we}, consisting of $1740$ examples focused on diverse knowledge granularity. Overall, we track the accuracy score averaged over these five evaluation datasets and two model trainings (PLM-3B and PLM-8B) during the data curation process. 

After the creation of our final \name{} dataset, we also evaluate them on \textit{five} more unseen evaluation datasets including \dynamath{} \cite{zou2024dynamath} and \lvista{} \cite{xiao2024logicvista} VL reasoning datasets containing $1000$ and $448$ examples, respectively. Further, we evaluate the visual understanding capabilities on \hbench{} \cite{guan2024hallusionbench}, a challenging perception dataset. In addition, we evaluate the general-purpose reasoning capabilities of a VLM on text-only reasoning datasets including math-centric \tmath{} \cite{lightman2023lets} and science-centric \tgpqa{} \cite{rein2024gpqa} containing $500$ and $198$ examples, respectively. To ensure consistency, we use \textit{accuracy} as the scoring metric and an identical prompt, instructing the model to always answer in \texttt{boxed} format, across all evaluations. Further, we use greedy decoding with maximum generation length of $2048$ across all the evaluation datasets. 
\begin{table}[t]
\centering
\resizebox{\textwidth}{!}{%
\begin{tabular}{lccccccc}
\toprule \textbf{Context Sources} & \textbf{Models}& \textbf{Average} & \textbf{MathVerse} & \textbf{MathVista} & \textbf{MathVision} & \textbf{MMMU-Pro} & \textbf{We-Math}\\
\toprule
\multirow{3}{*}{\virl{}\textsuperscript{\dag} \citep{vl-rethinker}}& 3B& 38.8& 32.9 & 58.6 & 25.7& 25.0& 52.0\\
& 8B& 41.3& 34.5 & 64.5 & 23.4& 28.2& 55.8\\
\rowcolor{blue!20}
\cellcolor{white}&Avg. & \textbf{40.1}& \textbf{33.7} & \textbf{61.6} & 24.5& \textbf{26.6}& \textbf{53.9}\\\hline
\multirow{3}{*}{\mathllava{} \citep{shi2024math}}& 3B& 36.3& 30.0 & 56.3 & 23.6& 21.7& 50.1\\
& 8B& 39.2& 33.7 & 62.2 & 23.4& 26.1& 50.5\\
 \rowcolor{blue!20}
\cellcolor{white}&Avg. & 37.7& 31.8 & 59.2 & 23.5& 23.9& 50.3\\\hline
\multirow{3}{*}{\ronevision{} \citep{yang2025r1}} & 3B& 35.7& 30.6 & 55.2 & 20.1& 22.8& 49.7\\
& 8B& 38.8& 32.5 & 60.5 & 22.0& 26.9& 52.2\\
\rowcolor{blue!20}
\cellcolor{white}& Avg. & 37.3& 31.6 & 57.9 & 21.1& 24.9& 51.0\\\hline
\multirow{3}{*}{\thinklite{} \textsuperscript{\dag} \citep{wang2025sota}}& 3B& 34.6& 25.5 & 53.7 & 25.3& 20.6& 48.1\\
& 8B& 39.5& 31.2 & 61.2 & 24.0& 27.7& 53.1\\
\rowcolor{blue!20}
\cellcolor{white} &Avg. & 37.1& 28.4 & 57.5 & \textbf{24.7}& 24.2& 50.6\\\hline
\multirow{3}{*}{\llavacot{} \citep{xu2024llava}}& 3B& 34.6& 28.5 & 54.2 & 19.0& 22.6& 48.6\\
& 8B& 37.9& 32.7 & 60.3 & 18.7& 26.5& 51.5\\
\rowcolor{blue!20}
\cellcolor{white}& Avg. & 36.3& 30.6 & 57.3 & 18.9& 24.5& 50.1\\\hline
\multirow{3}{*}{\mmk{} \citep{meng2025mm}} & 3B& 34.6& 28.6 & 58.1 & 19.0& 26.4& 40.6\\
& 8B & 37.3& 29.2 & 55.1 & 24.3& 26.4& 51.5\\
\rowcolor{blue!20}
\cellcolor{white}& Avg. & 36.0& 28.9 & 56.6 & 21.7& 26.4& 46.1 \\
\bottomrule
\end{tabular}%
}
\caption{\small{\textbf{Ranking the quality of the context from diverse datasets.} We train PLM-3B and PLM-8B on CoTs generated on the context (image, question) pairs from diverse dataset. Then, we rank them based on the average performance of these models on several VL reasoning downstream tasks. We mark the datasets which benefit from final answer correctness filtering in these experiments by \textsuperscript{\dag}.}}
\label{tab:ranked_context_sources}
\end{table}

\begin{table}[t]
\centering
\begin{tabular}{cccccccc}
\toprule \textbf{Mixing Sources} & \textbf{Models}& \textbf{Average} & \textbf{MathVerse} & \textbf{MathVista} & \textbf{MathVision} & \textbf{MMMU-Pro} & \textbf{We-Math}\\
\toprule
\multirow{3}{*}{Top 1} & 3B& 38.8& 32.9 & 58.6 & 25.7& 25.0& 52.0\\
 & 8B& 41.3& 34.5 & 64.5 & 23.4& 28.2& 55.8\\
\rowcolor{blue!20}
\cellcolor{white}& Avg. & \textbf{40.1}& 33.7 & 61.6 & 24.5& 26.6& 53.9\\\hline

\multirow{3}{*}{Top 2} & 3B& 37.8& 33.4&58.9&21.1&25.6&50.2\\
 & 8B& 39.3& 32.7&60.9&20.7&29.2&53.0\\
\rowcolor{blue!20}
\cellcolor{white} & Avg. & 38.6& 33.1&59.9&20.9&27.4&51.6\\\hline
\multirow{3}{*}{Top 4} & 3B& 36.7& 31.5&57.6&20.1&24.0&50.5\\
 & 8B& 39.7& 34.9&61.6&23.4&28.2&50.3\\
\rowcolor{blue!20}
\cellcolor{white}& Avg. & 38.2& 33.2&59.6&21.7&26.1&50.4\\\hline
\multirow{3}{*}{All} & 3B& 36.5& 33.1&58.3&17.8&23.2&49.9\\
 & 8B& 40.5& 35.8&61.7&25.3&28.2&51.3\\
 \rowcolor{blue!20}
\cellcolor{white} & Avg. & 38.5& 34.5&60.0&21.5&25.7&50.6\\
\bottomrule
\end{tabular}
\caption{\small{\textbf{Impact of mixing source datasets.} We mix the VL reasoning CoT data from diverse datasets to assess whether it leads to better performance than the individual datasets itself. We find that the individual dataset source, \virl{}, achieves the best performance.}}
\label{tab:source_mixing}
\end{table}

\begin{table}[t]
\centering
\resizebox{\textwidth}{!}{%
\begin{tabular}{ccccccccccc}
\toprule
\textbf{Method} & \textbf{Skill} & \textbf{Type}&\textbf{Models} & \textbf{Average} & \textbf{MathVerse} & \textbf{MathVista} & \textbf{MathVision} & \textbf{MMMU-Pro} & \textbf{We-Math}\\\toprule
\multirow{3}{*}{Original Data}& \multirow{3}{*}{-} & \multirow{3}{*}{-}& 3B&  38.8& 32.9 & 58.6 & 25.7  & 25.0& 52.0 \\
&&&8B & 41.8 & 35.3 & 65.8 & 22.4& 29.8& 55.6\\ 
& &&\cellcolor{blue!20}Avg.& \cellcolor{blue!20}40.1& \cellcolor{blue!20}33.7 & \cellcolor{blue!20}61.6 & \cellcolor{blue!20}24.5  & \cellcolor{blue!20}26.6& \cellcolor{blue!20}53.9\\\hline
\multirow{3}{*}{\textbf{Caption and Solve}}& \multirow{3}{*}{\makecell{Perception\\(Auxiliary Signal)}}& \multirow{3}{*}{Augment} & 3B& 39.7& 35.7 & 56.6 & 24.7  & 26.7& 55.0\\
&&& 8B& 43.0& 38.3 & 63.2 & 26.3  & 30.4& 56.9\\
&& & \cellcolor{blue!20} Avg. & \cellcolor{blue!20}41.4 (\textcolor{blue}{+3.3pp})& \cellcolor{blue!20}37.0 &\cellcolor{blue!20} 59.9 & \cellcolor{blue!20}25.5  & \cellcolor{blue!20}28.6& \cellcolor{blue!20}56.0\\\hline
\multirow{3}{*}{\makecell{Visual Perturb}}& \multirow{3}{*}{\makecell{Perception\\(Robustness)}} & \multirow{3}{*}{Replace}&3B& 36.5& 31.2 & 57.5 & 18.1  & 25.8& 49.9 \\
&&& 8B & 40.5& 31.9 & 63.4 & 22.7  & 29.5& 54.8 \\
&&&\cellcolor{blue!20}Avg.&\cellcolor{blue!20}38.5& \cellcolor{blue!20}31.6 & \cellcolor{blue!20}60.5 & \cellcolor{blue!20}20.4  & \cellcolor{blue!20}27.7& \cellcolor{blue!20}52.4\\\hline
\multirow{3}{*}{\makecell{Text-Rich Images}} & \multirow{3}{*}{\makecell{Perception\\(Synthetic Images)}} & \multirow{3}{*}{Replace}& 3B& 37.1& 30.2 & 55.5 & 22.7  & 27.1& 49.9 \\
&&& 8B& 40.5& 35.3 & 61.0 & 21.1  & 30.6& 54.5 \\
&&&\cellcolor{blue!20}Avg. &\cellcolor{blue!20}38.8 & \cellcolor{blue!20}32.8 & \cellcolor{blue!20}58.3 &\cellcolor{blue!20} 21.9  & \cellcolor{blue!20}28.9&\cellcolor{blue!20} 52.2\\\hline
\multirow{3}{*}{Perceptual Redundancy}& \multirow{3}{*}{\makecell{Perception\\(Feasibility wo/ image)}} & \multirow{3}{*}{Filter}& 3B & 36.0& 31.2 & 54.6 & 21.4  & 24.6& 48.4\\
&&&8B & 36.9& 32.7 & 58.0 & 18.1  & 27.8& 48.1 \\
&&&\cellcolor{blue!20} Avg. & \cellcolor{blue!20}36.5& \cellcolor{blue!20}32.0 & \cellcolor{blue!20}56.3 & \cellcolor{blue!20}19.8  & \cellcolor{blue!20}26.2& \cellcolor{blue!20}48.3 \\\hline
\multirow{3}{*}{Shallow Perception} & \multirow{3}{*}{\makecell{Perception\\(Feasibility w/ caption)}} & \multirow{3}{*}{Filter}& 3B & 34.7& 29.3 & 53.4 & 20.7  & 23.8& 46.2 \\
&&&8B &  36.5& 29.2 & 57.3 & 21.4  & 28.1& 46.3\\
&&&\cellcolor{blue!20}Avg. &\cellcolor{blue!20}35.6 & \cellcolor{blue!20}29.3 &\cellcolor{blue!20} 55.4 &\cellcolor{blue!20} 21.1  &\cellcolor{blue!20} 26.0& \cellcolor{blue!20}46.3\\\hline
\multirow{3}{*}{\textbf{Text-Only Data}}& \multirow{3}{*}{\makecell{Problem-Solving\\(Cross-modal transfer)}}& \multirow{3}{*}{Augment} & 3B& 41.9& 38.1 & 60.1 & 25.7  & 26.8& 58.9\\
&&& 8B& 44.2& 41.2 & 63.4 & 27.3  & 31.4& 57.6\\
&&&\cellcolor{blue!20}Avg. &\cellcolor{blue!20}43.1 (\textcolor{blue}{+7.5pp})& \cellcolor{blue!20}39.7 & \cellcolor{blue!20}61.8 & \cellcolor{blue!20}26.5  & \cellcolor{blue!20}29.1& \cellcolor{blue!20}58.3 \\\hline
\multirow{3}{*}{\makecell{Increased Distractors}} & \multirow{3}{*}{\makecell{Problem-Solving\\(Robustness)}} & \multirow{3}{*}{Replace} &3B& 33.1& 23.4 & 51.3 & 19.7  & 22.3& 48.8\\
&&& 8B& 36.2& 27.4 & 53.2 & 20.1  & 27.5& 52.6 \\
&&&\cellcolor{blue!20}Avg.& \cellcolor{blue!20}34.6& \cellcolor{blue!20}25.4 & \cellcolor{blue!20}52.3 &\cellcolor{blue!20}19.9  &\cellcolor{blue!20} 24.9& \cellcolor{blue!20}50.7 \\\hline
\multirow{3}{*}{Length}& \multirow{3}{*}{\makecell{Problem-Solving\\(Long Thinking)}} & \multirow{3}{*}{Filter}&3B & 33.9& 28.3 & 50.3 & 19.7  & 24.6& 46.8\\
&&& 8B & 38.8& 32.6 & 60.7 & 19.4  & 28.3& 53.2\\
&&&\cellcolor{blue!20}Avg. &\cellcolor{blue!20}36.4& \cellcolor{blue!20}30.5 &\cellcolor{blue!20} 55.5 &\cellcolor{blue!20} 19.6  & \cellcolor{blue!20}26.5& \cellcolor{blue!20}50.0\\\hline
\multirow{3}{*}{Uniform Difficulty} & \multirow{3}{*}{\makecell{Problem-Solving\\(Hardness)}}& \multirow{3}{*}{Filter}& 3B& 33.1& 25.1 & 52.9 & 17.4  & 24.1& 45.8\\
&&&8B& 36.1& 28.6 & 56.8 & 21.4  & 24.9& 48.8\\
&&&\cellcolor{blue!20}Avg. & \cellcolor{blue!20}34.6& \cellcolor{blue!20}26.9 & \cellcolor{blue!20}54.9 & \cellcolor{blue!20}19.4  & \cellcolor{blue!20}24.5& \cellcolor{blue!20}47.3\\ \bottomrule
\end{tabular}%
}
\caption{\small{\textbf{Results for data intervention experiments.} We compare the performance of VL reasoners trained on datasets created using diverse data intervention strategies. Our results show that achieving better performance than the original data, which is obtained by selecting the right data sources and their mixtures, is non-trivial. We also report the results for the best-performing configuration from each intervention strategy. We find that augmenting the original CoT with image captions (\textit{caption and solve}), and mixing with \textit{text-only reasoning} data reliably improve VL reasoning performance.}}
\label{tab:data_intervention_results}
\end{table}

\section{Experiments}
\label{sec:experiments}

\subsection{Impact of Context Curation}
\label{sec:context_curation_results}

\paragraph{\textbf{Sourcing.}} We present results for training PLM-3B and PLM-8B on reasoning data constructed using diverse context sources in Table \ref{tab:ranked_context_sources}. Specifically, we compute the average performance across five VL reasoning datasets and rank the context sources from highest to lowest accuracy. For each context source, we report the best results achieved from one of two scenarios: unfiltered CoT data and filtered CoT data based on final answer correctness.\footnote{We find that \virl{} and \thinklite{} benefit from the final answer correctness filter, while the others perform better with the unfiltered dataset (Appendix Table \ref{apptab:filtering_or_no_filtering}).} Our experiments reveal that the choice of context source has a significant impact on downstream VL reasoning performance. Specifically, we observe a relative gap of $11.4$ percentage points (pp) between the average performance of the lowest (\mmk{}, $36.0\%$) and highest performing source datasets (\virl{}, $40.1\%$). This highlights the choice of context source has a significant impact on downstream VL reasoning performance.

\paragraph{\textbf{Mixing.}}We next assess the impact of mixing contexts from different data sources. To this end, we combine examples from the previous stage for the top-2, top-4, and all source datasets, and select a random subset of size $50$K. The results of training PLM-3B and PLM-8B on these mixtures are presented in Table \ref{tab:source_mixing}. Interestingly, we find that mixing data from diverse sources does not improve the performance over the best-performing dataset, \virl{}. This suggests that mixing instances from diverse sources can degrade the performance on the VL reasoning tasks.\footnote{We leave the exploration of complex data mixing strategies \cite{Xie2023DoReMiOD} for VL reasoning as future work.}

\subsection{Impact of Data Interventions}
\label{sec:data_interventions}

Starting from the best data from the previous stage, we note the impact of several data intervention experiments on VL reasoning capabilities in Table \ref{tab:data_intervention_results}. Specifically, we target the strategies at improving the perception and problem-solving capabilities of the VLMs.\footnote{We present the results for the best-performing configuration for the interventions. For instance, we experiment with diverse ways of implementing \textit{caption and solve} method, and show the results for the best-performing variant.} 

\begin{wraptable}[15]{r}{0.4\textwidth}
\small
\setlength{\tabcolsep}{4pt}
\centering
\begin{tabular}{lccc}
\toprule
\textbf{Data} & \textbf{Avg.} & \textbf{PLM-3B} & \textbf{PLM-8B} \\
\midrule
Baseline & 32.2& 30.4 & 34.0\\\hline
VL& 39.2& 34.6 & 43.8 \\
Text& 62.7& 60.2 & 65.2 \\
VL + Text & 59.7& 57.0& 62.4\\
\bottomrule
\end{tabular}
\caption{\small{\textbf{Generalization to MATH500.} We present the performance of VL reasoners on text-only MATH500 reasoning task. The baseline refers to the base models performance, VL refers to exposure to just VL reasoning data, Text refers to training with re-annotated \ot{}, and VL+Text refers to mixture of VL and text reasoning CoT data.}}
\label{tab:mix_vl_text_only}
\end{wraptable}

We find that the best-performing data intervention for enhanced perception is the \textit{caption and solve} strategy, which provides auxiliary visual signals about the image description in the reasoning CoT data (i.e., \texttt{<caption> description </caption> solution \boxed{answer}}). Specifically, this approach improves accuracy averaged across multiple downstream tasks and model training runs by $3.3$pp. In addition to performance gains, we show that the ability to generate image description before problem-solving offers novel efficiency improvements in test-time scaling scenarios, such as generating multiple solutions per (image, question) pair (\S\ref{sec:efficiency_tts}). 

In addition, we observe that the best-performing data intervention for enhanced problem-solving is the inclusion of \textit{text-only reasoning} data in the training mixture. In particular, we observe a relative accuracy improvement of $7.5$pp over the original dataset across multiple VL reasoning tasks and training runs. Apart from improving the VL reasoning, we want our VL reasoners to be general-purpose reasoners, thus, we study the performance of VLMs on blind (text-only) reasoning evaluation dataset \tmath{} in Table \ref{tab:mix_vl_text_only}. 

Specifically, we find that the base models PLM-3B and PLM-8B do not perform well on the \tmath by achieving an accuracy of $32.2\%$ and $34.0\%$, respectively. We find that training these models with VL reasoning CoT data (baseline data in Table \ref{tab:data_intervention_results}) improves the average performance to $39.2\%$, indicating cross-modal reasoning transfer (vision-language to language-only). However, this performance still lags behind training the VLMs with text-only reasoning data (\ot{}). But, training with the mixture of VL and text-only reasoning data improves the accuracy from $39.2$ to $59.7$ while improving the VL reasoning performance too. Thus, we will apply \textit{caption and solve} and \textit{text-only reasoning} data intervention strategies to create our final VL reasoning data. Interestingly, we find that most methods lead to poorer performance than the baseline, which suggests that it is non-trivial to improve over the best-chosen data source using the targeted data interventions.

\subsection{Scaling to Million Samples}
\label{sec:scaling_million}

\begin{figure}[t]
\centering
\begin{subfigure}[ht]{0.32\textwidth}
 \centering
 \includegraphics[width=\textwidth]{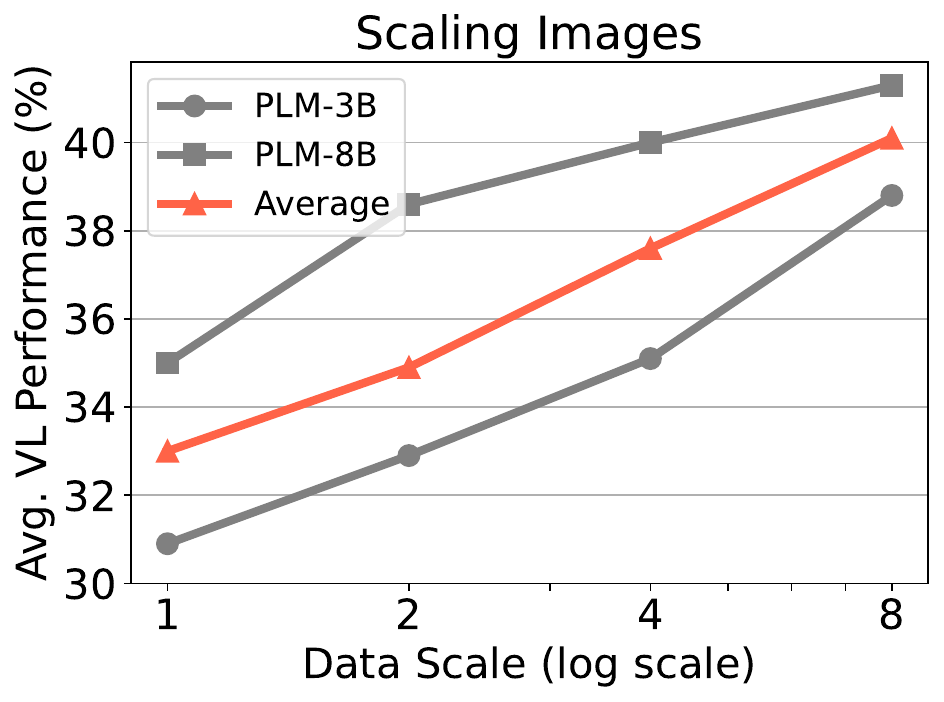}
 \caption{}
 \label{fig:scaling_images}
\end{subfigure}%
\begin{subfigure}[ht]{0.32\textwidth}
 \centering
 \includegraphics[width=\textwidth]{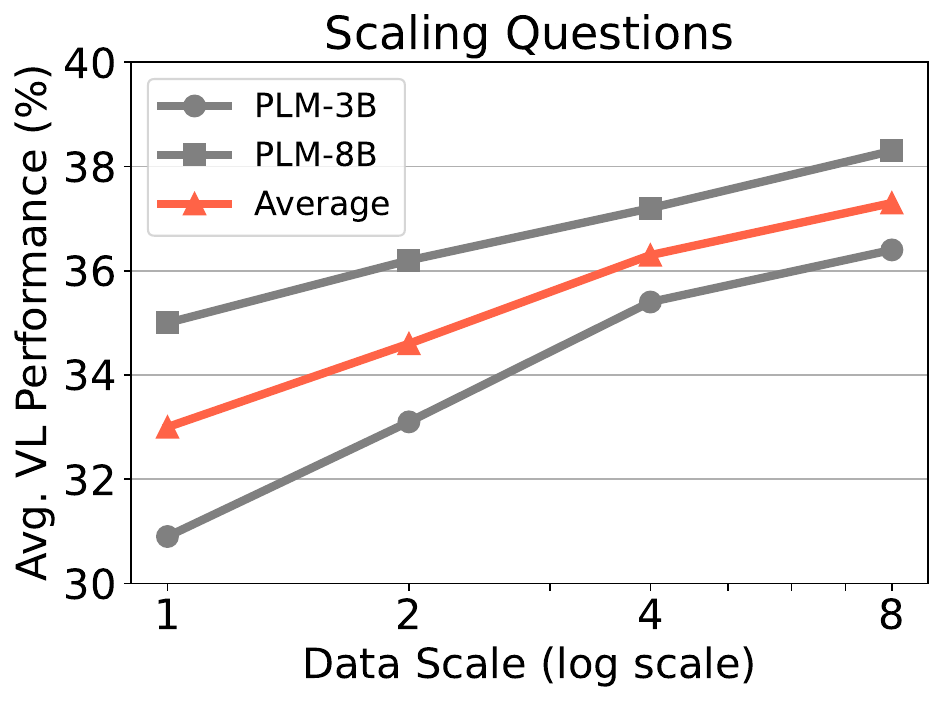}
 \caption{}
 \label{fig:scaling_questions}
\end{subfigure}
\begin{subfigure}[ht]{0.32\textwidth}
 \centering
 \includegraphics[width=\textwidth]{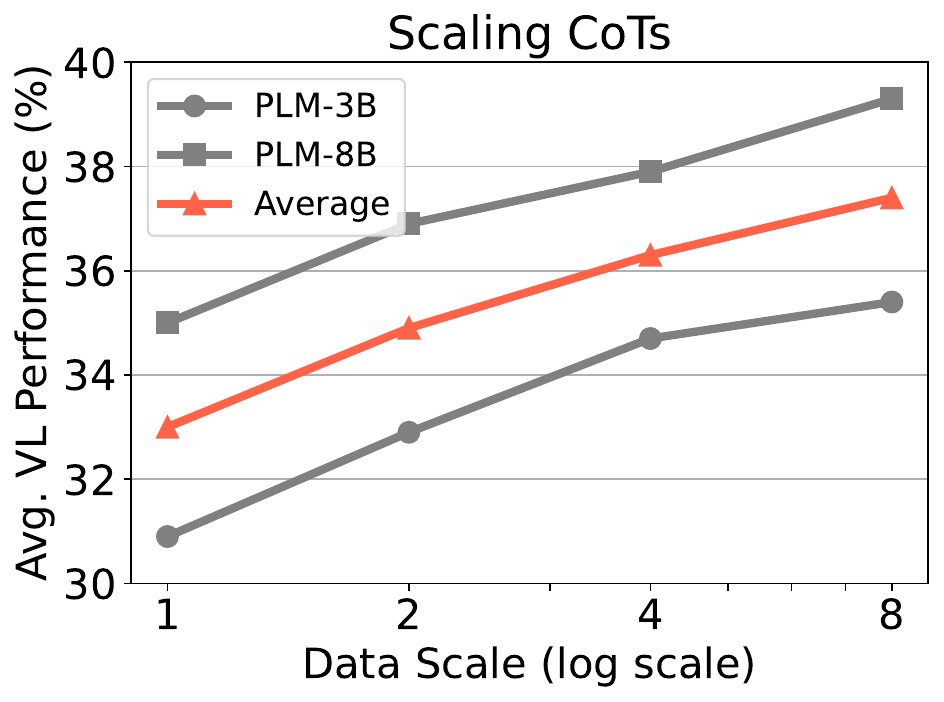}
 \caption{} 
 \label{fig:scaling_responses}
\end{subfigure}
\caption{\small{\textbf{Impact of scaling diverse data axes in VL reasoning data.} We train PLM-3B and PLM-8B on datasets of varying sizes (setup explained in \S \ref{sec:scaling_diverse_modalities}). The results show that the reasoning performance consistently improves as we scale each data axis: (a) images, (b) synthetic questions per image, and (c) CoTs per (image, question) pair.}}
\label{fig:scaling_modalities}
\end{figure}

\paragraph{\textbf{Scaling Trends Across Data Axes.}} To synthesize large-scale data, we study the scaling behavior of the best-performing dataset, \virl{}, including the number of unique images, the number of questions per image, and the number of CoTs per (image, question) pair. We present the results for accuracy averaged across the validation downstream datasets in Figure \ref{fig:scaling_modalities}. Interestingly, we find that the performance of the VL reasoners improves with scaling in images, new questions, and CoTs across model trainings of both PLM-3B and PLM-8B.

\begin{figure}[t]
\centering
\begin{subtable}[t]{0.4\textwidth}
\centering
\begin{tabular}{lc}
\toprule
\textbf{Statistic}& \textbf{Number (in K)} \\\toprule
Total instances& 2480\\
Number VL instances & 1440\\
Number Text-Only instances& 1040\\
Number Unique images & 28 \\
Number Unique questions& 350\\
Avg. question length (words) & 0.057\\
Avg. CoT length (words)&0.601\\\bottomrule
\end{tabular}
\caption{}
\label{fig:dataset_statistics}
\end{subtable}
\hspace{1cm}
\begin{subtable}[t]{0.4\textwidth}
\centering
\begin{tabular}{lcc}
\toprule
\textbf{KeyWord}& \textbf{Category} & \textbf{Number (in K)} \\\toprule
\texttt{Identify} & Perception & 1950\\
\texttt{Recall} & Perception & 1127\\
\texttt{Provided} & Comprehension & 1040\\
\texttt{First}& Planning & 2388\\
\texttt{Wait}& Reflection & 32\\\bottomrule
\end{tabular}
\caption{}
\label{fig:keyword_analysis}
\end{subtable}
\caption{\small{\textbf{\name{} statistics.} (a) We present key statistics of the \name{} dataset. Notably, \name{} is a large-scale dataset with $2.5$M instances with $350$K unique questions. (b) We also provide statistics on several keywords in the \name{} CoTs that elicit diverse reasoning behaviors, such as perceptual understanding, question comprehension, planning, and reflection.}}
\label{fig:stats_keyword_analysis}
\end{figure}

\begin{figure}[h]
\centering
\includegraphics[width=\linewidth]{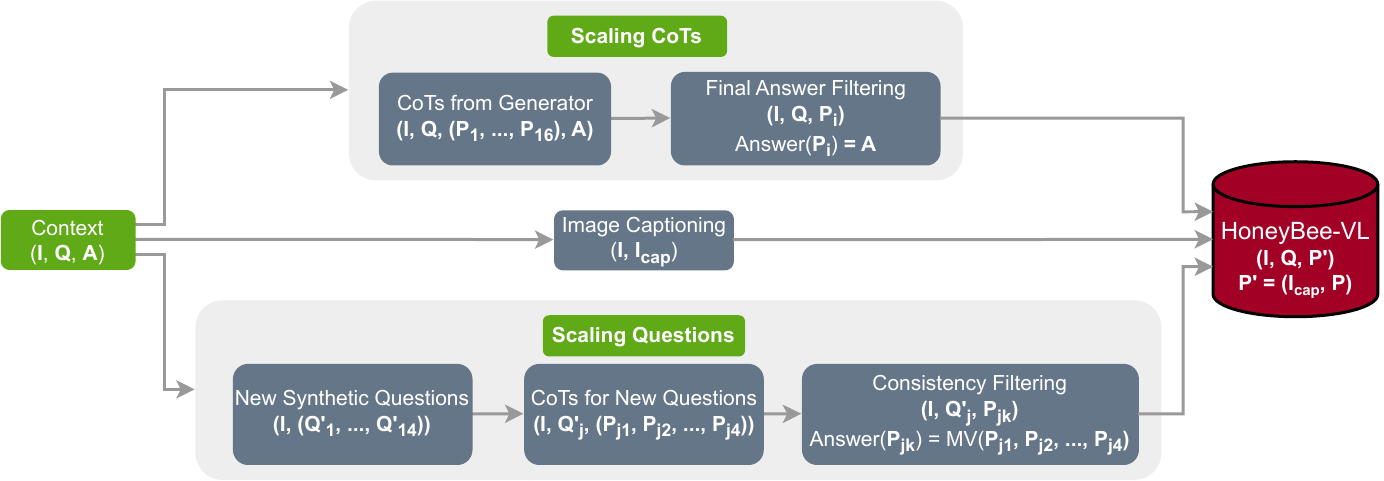}
\caption{\small{\textbf{\name{}-VL scaling pipeline.} We present the pipeline to scale the number CoTs per (real image, question) pair (\textbf{top}), and scale the number of questions per real image (\textbf{bottom}) from the context datasets. We denote majority voting over the several predicted answers as MV. Ultimately, this pipeline leads to the creation of $1.5$M VL reasoning instances.}}
\label{fig:scaling_data_pipeline}
\end{figure}

\paragraph{\textbf{Putting Everything Together.}} Firstly, we include all the real data in the \virl{} datasets, which consists $39$K (image, question, and final answer) tuples. Since the amount of real data is limited, we scale the data by generating several CoTs for all (image, question) pairs, and synthetically creating new questions (Figure \ref{fig:scaling_data_pipeline}). Specifically, we generate $16$ CoTs per real image and existing question pair (\textit{scaling CoTs}). To retain the highest quality data, we filter out CoTs that do not lead to the correct final answer, leading to roughly $400$K instances. To scale questions, we $14$ new questions per image, resulting in $15$ questions per image in total (\textit{scaling questions}). However, we do not have access to the final answers when generating new questions. To address this, we generate $4$ CoTs per new question and use majority voting (an answer occurring three or more times) as a proxy for the final answer \cite{prasad2024self}. We then retain the CoTs for which the predicted answer matches the proxy final answer in the \textit{scaling questions} scenario, leading to roughly $1$M instances. In parallel, we generate image captions for all images in the source datasets. Following the \textit{caption and solve} strategy, we combine the (image, question, solution CoT) tuples from the scaling CoTs and questions pipelines with the image captions to construct (image, question, (image caption, solution CoT)) tuples. This process results in the construction of the VL subset of the \name{} dataset, consisting of \textbf{1.5M} instances. We then merge this subset with the text-only reasoning data, consisting of \textbf{1M} instances, to obtain a high-quality, large-scale final \name{} dataset of size \textbf{2.5M}.

\paragraph{\textbf{Dataset Statistics.}} We present the dataset statistics in Figure \ref{fig:dataset_statistics}. Beyond the total dataset size, we highlight that \name{} contains $28$K unique images and $350$K unique questions. The average length of CoTs (image caption and solution CoT combined) is approximately $600$ words ($780$ tokens). Following \cite{deng2025openvlthinker}, we analyze the occurrence of several keywords in our CoTs that are crucial for diverse reasoning behaviors, including visual understanding (perception), question comprehension, planning, and reflection. This suggests that the reasoning CoTs in our dataset encourage VL reasoners to elicit more complex reasoning (such as reflection) behaviors than existing VL reasoning datasets.

\subsection{Training VL Reasoners with \name{}}
\label{sec:training_with_honeybee}

\begin{wrapfigure}{R}{0.4\textwidth}
\centering
\includegraphics[width=0.38\textwidth]{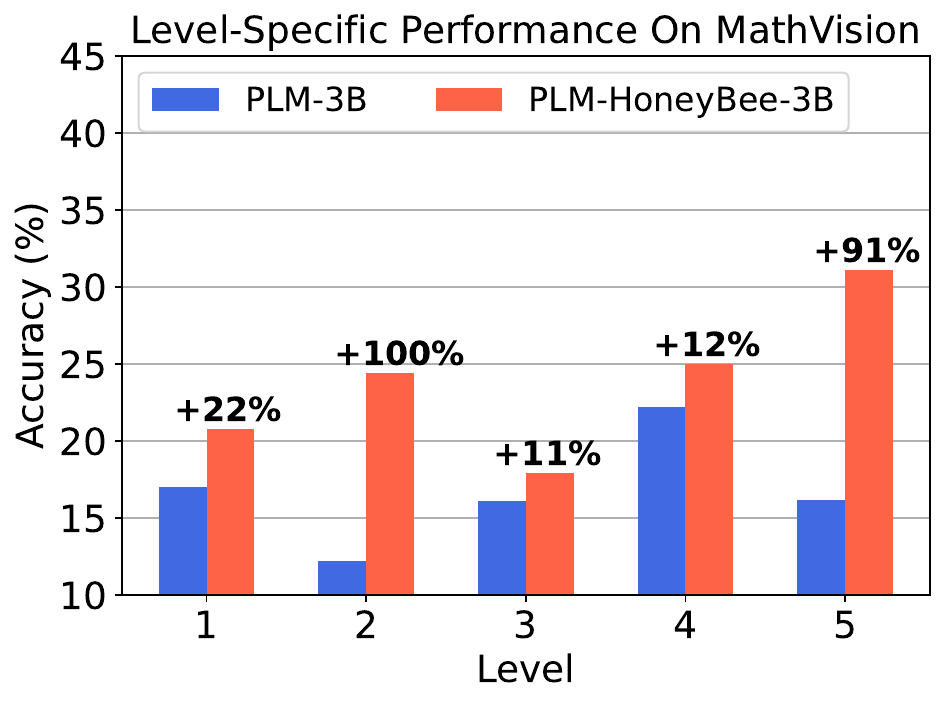}
\caption{PLM-3B trained on \name{} outperforms the base model across all \mvision{} difficulty levels.}
\label{fig:fine_grained}
\end{wrapfigure}

\paragraph{\textbf{\name{} elicits strong reasoning.}} Here, we assess the performance of PLMs of different sizes ($1$B, $3$B) trained on the entire $2.5$M \name{} dataset. We also compare them against several high-performing VLMs capable of generating CoTs to solve VL reasoning tasks. Results on several evaluation datasets are presented in Table \ref{tab:main_table}. We find that VL reasoners trained with \name{} achieve the highest average accuracy across all model size categories. In particular, PLM-\name{}-1B achieves a relative performance improvement of $28$pp over \internthree{}-1B-Instruct. Furthermore, PLM-\name{}-3B and PLM-\name{}-8B achieve relative gains of $8.4$pp and $2.7$pp over \qwen{}-3B-Instruct and \qwen{}-8B-Instruct, respectively. Moreover, we perform a fine-grained comparison between the base PLM-3B and PLM-\name{}-3B across diverse difficulty levels in the \mvision{} dataset (Figure \ref{fig:fine_grained}). We find that \name{} improves performance across all difficulty levels, reaching up to $100$pp relative gains for level 2. This highlights that \name{} can be used to enhance reasoning capabilities at various difficulty levels. Overall, we show that our high-quality curated data can outperform existing SOTA methods that provide little public information about their reasoning CoT data. In addition, we observe that models trained on \name{} generalize well and achieve best-in-class performance on evaluation datasets that were unseen during curation, including robust reasoning in \dynamath{}, logical reasoning \lvista{}, perception-centric \hbench{}, and STEM-reasoning \tgpqa{}.

\paragraph{\textbf{\name{} data scaling.}} To understand the data scaling behavior with \name{}, we train PLM models on various subsets ($50$K, $250$K, $2.5$M) of the data. Results averaged across the five VL evaluation datasets are shown in Figure \ref{fig:summary_results}. We observe that the performance of \name{}-trained PLMs, across all sizes, continues to improve as the dataset size increases. In fact, we find that performance has not saturated even at the $2.5$M scale. This suggests that, with sufficient training budget, one could further scale the size of \name{} to achieve additional performance improvements. We show the scaling results across individual datasets in Appendix Figure \ref{fig:scaling_per_dataset}. Further, we compare the PLM-3B trained with the \name{} data and existing CoT VL reasoning datasets in Figure \ref{fig:comparison_with_existing}, and show massive relative gains upto $39$pp across the five VL reasoning datasets. Apart from this, we use a small subset of our data to finetune the teacher model itself, reminiscent of self-improvement \cite{singh2023beyond}, and show that \name{} data can be used to achieve reasoning performance from the generator model too (Appendix \ref{app:self_improvement_experiment}).

\paragraph{\textbf{RL training of \name{} model.}} Supervised finetuning with CoT data serves as a warm start for RL training, enabling further improvements in reasoning capabilities \cite{liu2025acereason,deng2025openvlthinker}. Thus, we perform RL training using the GRPO algorithm \cite{shao2024deepseekmath} on top of the PLM-\name{}-3B model with VL verifiable data \cite{mmopenr1}. We present results on diverse VL reasoning tasks and compare them to those of a supervised and RL-tuned VL reasoner, OpenVLThinker-v1.2-3B, in Table \ref{tab:rl_main_results}. We find that RL training on top of \name{}-SFT models outperforms OpenVLThinker-v1.2-3B, with a $9.2$pp relative gain in average accuracy. In particular, the RL-trained model achieves substantial improvements on the \mvision{} and \wemath{} datasets compared to the SFT model, with relative margins of $27.5$pp and $10.0$pp, respectively.

\begin{table}[t]
\centering
\resizebox{\textwidth}{!}{%
\begin{tabular}{ccccccc}
\toprule \textbf{Algorithm} & \textbf{Average} & \textbf{MathVerse} & \textbf{MathVista} & \textbf{MathVision} & \textbf{MMMU-Pro} & \textbf{We-Math}\\\toprule
OpenVLThinker-v1.2-3B \cite{deng2025openvlthinker} & 42.3 & 34.1&	63.9	&25.0	&\textbf{28.7}	&59.8\\\hline
PLM-\name{}-3B & 44.3 & 42.8	&61.2&	\textbf{29.9}	&28.4&	59.3\\
\rowcolor{blue!20}PLM-\name{}-3B-GRPO & \textbf{46.2}& \textbf{43.5}	&\textbf{64.9}&	28.4	&28.3&	\textbf{65.8} \\\bottomrule
\end{tabular}%
}
\caption{\small{\textbf{RL training on top of \name{} trained VLM.} We present results for training PLM-3B with \name{} data using supervised finetuning, followed by a round of RL training with the GRPO algorithm on a verifiable VL dataset. Furthermore, we compare its performance to that of a state-of-the-art RL-tuned VL reasoner, OpenVLThinker-v1.2-3B, which has a similar model capacity.}}
\label{tab:rl_main_results}
\end{table}

\begin{figure}[t]
\centering
\begin{subfigure}[h]{0.60\textwidth}
\centering
\includegraphics[width=\textwidth]{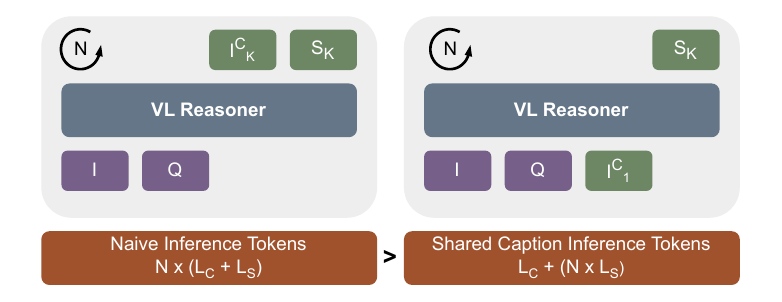}
\caption{\small{}}
\label{fig:efficiency_method}
\end{subfigure}
\begin{subfigure}[h]{0.35\textwidth}
\centering
\includegraphics[width=\textwidth]{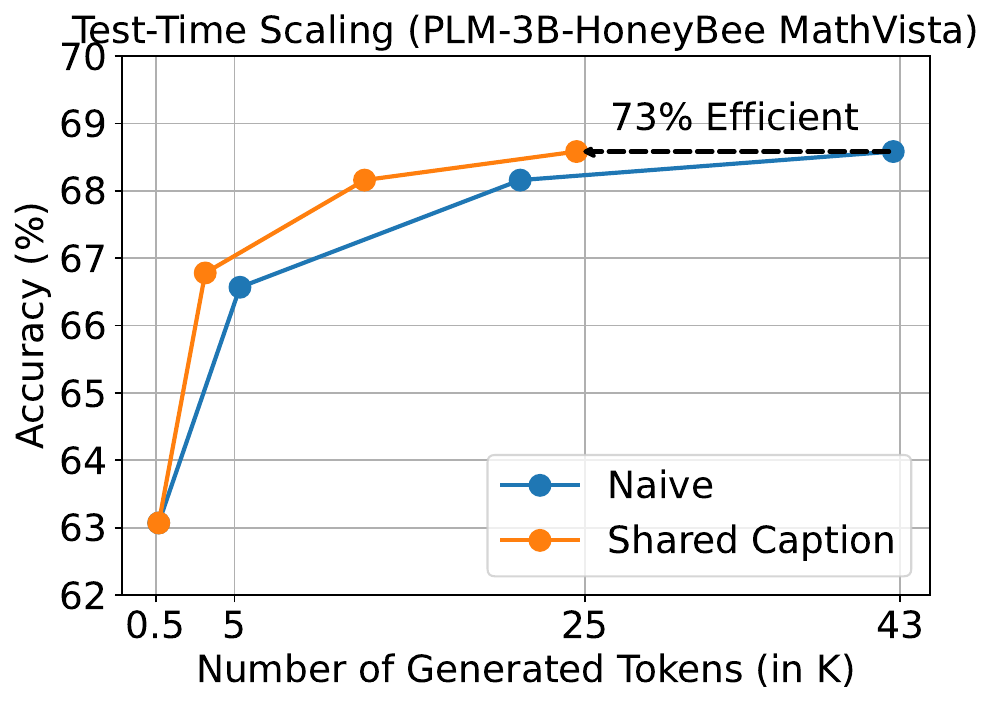}
\caption{\small{}}
\label{fig:efficiency_results}
\end{subfigure}
\caption{\small{\textbf{Shared caption decoding for efficient test-time scaling (TTS).} (a) We illustrate the naive approach to TTS in VL reasoning and compare it to the proposed decoding strategy. After training with the \name{} dataset, we show that the number of generated tokens can be significantly reduced by reusing the caption generated for a given image in subsequent problem-solving attempts. We note that $I^C_1$ is the image caption generated in the first attempt of solution generation. (b) We present the accuracy trend as a function of the number of generated tokens (number of solution attempts up till $64$) for PLM-3B trained with \name{} on the downstream \mvista{} dataset.}}
\label{fig:efficiency_pipeline}
\end{figure}

\subsection{Efficient test-time scaling with Shared captions}
\label{sec:efficiency_tts}

We observe that each CoT in the \name{} dataset has two parts: an understanding component ($I^C$, image caption tokens) and a problem-solving component ($S$, solution tokens), so $C = [I^C; S]$. In test-time scaling (TTS) methods like self-consistency \cite{wang2022self}, we generate $N > 1$ CoTs for a reasoning problem $(I, Q)$ and use majority voting over answers. The naive TTS approach generates the full CoT $N$ times. Instead, we propose \textit{shared captions}: generate the full CoT once as $(I^C_1, S_1)$, then reuse $I^C_1$ as context for subsequent solution generations $S_K$ (Figure \ref{fig:efficiency_method}). This reduces number of generated tokens and, since inference FLOPs scale with token count \cite{kaplan2020scaling}, improves inference efficiency. We empirically test this with PLM-3B trained on \name{} using \mvista{} (Figure \ref{fig:efficiency_results}), generating $N=64$ solutions at temperature $0.7$ per (image, question) pair. The naive approach produces $42.6$K tokens ($671$ per attempt), while \textit{shared captions} achieves similar performance with only $24.5$K tokens ($280$ captioning, $390$ problem-solving per attempt), a $73\%$ reduction in tokens and FLOPs. Thus, \name{} enables inference-time efficiency and strong VL reasoning.

\begin{figure}[t]
    \centering
    \includegraphics[width=0.7\linewidth]{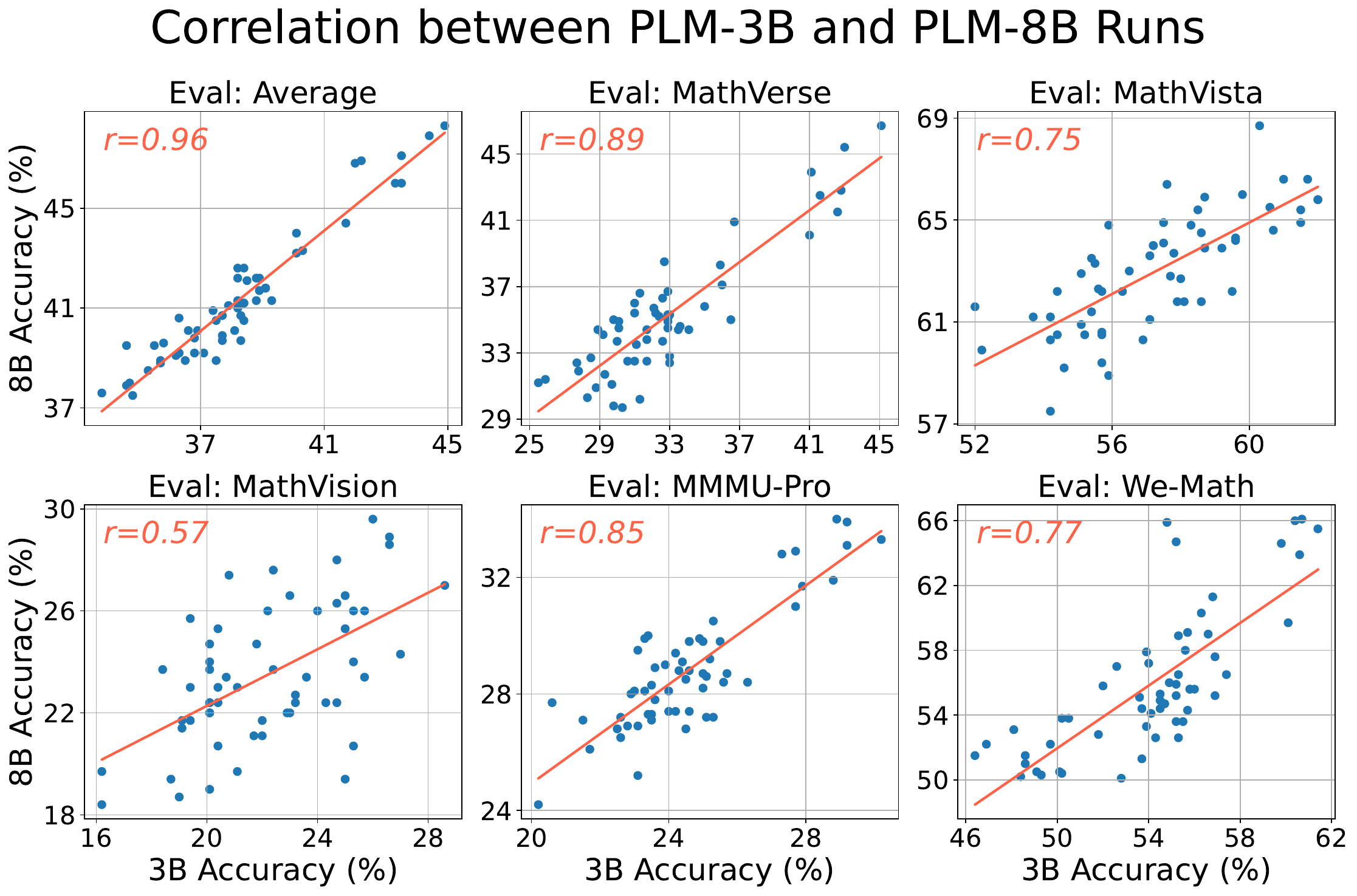}
    \caption{\small{\textbf{Model Correlation Analysis.} We present the correlation between the performance of PLM-8B and PLM-3B, each trained with various data distributions throughout this work, across diverse downstream evaluation tasks. A higher correlation implies greater predictability between model performances; that is, a data curation method that is optimal for a $3$B VLM is more likely to be optimal for an $8$B VLM.}}
    \label{fig:model_correlation}
\end{figure}

\subsection{Correlation Analysis}
\label{sec:analysis}

For our data curation, we train PLM-3B and PLM-8B and evaluate them on five VL reasoning tasks for each experiment to make robust decisions that genuinely improve reasoning performance across multiple model scales and evaluations. Here, we aim to understand the correlation between the performance achieved by PLM-8B and PLM-3B on average (and individual dataset) accuracy across all experiments conducted in this work (Figure \ref{fig:model_correlation}), and observe a strong correlation in average accuracy. This indicates that data curation methods have similar impacts on average accuracy across different model scales. Similar to \cite{magnusson2025datadecide} for LLM pretraining, a broader implication of this analysis is that future research can perform optimal data selection strategies with a small VLM, and it is likely to be optimal for a larger VLM.
\section{Related Work}

\paragraph{\textbf{Chain-of-Thought Reasoning Datasets}.} Prior work \citep{wei2022chain,kojima2022large} has shown that large language models (LLMs) can solve diverse reasoning tasks (e.g., math and logic) by generating chain-of-thoughts (CoTs) before providing their final answers at inference. This approach allows models to utilize additional computation at test time and further improve performance by aggregating decisions across several reasoning CoTs using methods such as self-consistency \citep{wang2022self}. Notably, STaR \citep{zelikman2022star} demonstrates that an LLM can be prompted to generate CoTs for novel questions, which can then be filtered based on final answer correctness. This synthetic process removes the need for expensive and time-consuming CoT collection from human annotators. Since then, several datasets \citep{li2024common,mitra2024orca,li2024numinamath,toshniwal2024openmathinstruct,toshniwal2024openmathinstruct2,bercovich2025llamanemotronefficientreasoningmodels} have been proposed that bootstrap math CoT reasoning traces from strong LLMs \citep{achiam2023gpt,guo2025deepseek}. In particular, OpenThoughts \citep{guha2025openthoughts} is a state-of-the-art CoT reasoning dataset, constructed through comprehensive ablation across several key data design decisions and ultimately scaled to $1.2$M reasoning examples, predominantly centered on math problem-solving. In comparison, our work focuses on the more complex scenario of math reasoning in visual contexts, where vision-language (VL) models must integrate information from multiple modalities, image and text, instead of just text. Specifically, we create \name{}, a high-quality VL CoT dataset that elucidates the data design process, including data sources, filtering strategies, and scaling properties at multiple model scales (e.g., 3B and 8B). Furthermore, we scale our dataset to $5$M VL instances and observe consistent improvements with increased data size.

\paragraph{\textbf{VL Reasoning Datasets.}} Math problem-solving serves as a critical testbed for assessing the reasoning capabilities of VL models, as it requires them to develop a deep understanding and interpretation skills across diverse visual categories such as geometry, functions, and charts \citep{lu2023mathvista,zhang2024mathverse,wang2024measuring}. To enable VL reasoning, prior work has proposed reasoning datasets, such as MathV360K \citep{shi2024math}, which selects and creates high-quality questions from a collection of existing image-question datasets. However, this dataset does not provide any CoT data, making it unsuitable for eliciting complex reasoning behaviors in VL models. Other datasets, such as MAVIS \citep{zhang2024mavis}, focus on creating novel questions and CoTs for plane geometry and functions using an automatic data engine. However, such synthetic data is unsuitable for achieving strong performance on diverse real-world visual scenarios, such as tabular understanding, IQ tests, science question-answering, and math reasoning on natural scenes. More recently, datasets such as \llavacot{}\citep{xu2024llava} and \ronevision{} \citep{yang2025r1} have sourced image-question data from diverse sources, including the web and textbooks, followed by CoT generation and multiple rounds of data curation from capable teacher VL models \citep{hurst2024gpt,team2023gemini}. However, it remains unclear whether the benefits of these datasets are attributable to differences in data sources, choice of teacher models, or specific data filtering strategies. In this work, we take a comprehensive approach by re-annotating the CoTs in each dataset with a given teacher model to understand the impact of diverse data sources, and we also study the effect of mixing the contexts in these datasets. Further, we curate a list of interventions (e.g., filtering, augmentation, and replacement) and test their impact on enhancing the performance of VL models of capacity. Subsequently, our approach examines the effect of scaling across several data dimensions, such as unique images, questions, and CoTs, which ultimately enables the development of \name{}, one of the largest VL reasoning datasets to date.

\paragraph{\textbf{VL Data Curation.}} Reasoning in visual contexts is a fundamental capability of modern AI systems \citep{2023GPT4VisionSC,team2024gemini,meta2025llama,team2025kimi,bai2025qwen25,chen2024expanding}, enabling a wide range of real-world applications such as visual data analysis and scientific discovery. However, the training data for these systems are not publicly available, making it challenging to study the underlying science behind the development of state-of-the-art VL reasoning CoT datasets. In this work, we aim to investigate the science behind curating high-quality, large-scale VL reasoning datasets. To this end, we curate a list of several data interventions designed to improve the perception and reasoning capabilities of these models. Prior work \citep{li2025vision} has shown that visual perturbations (e.g., rotation, distractors) can enhance perceptual robustness and lead to better mathematical reasoning. In addition, \cite{zhang2024mathverse} highlights that VL models often rely primarily on the information about the visual content provided in the question, rather than truly integrating visual information. Further, \cite{yue2024mmmu} argues that augmenting the original questions to have 10 options instead of 4 reduces VL model performance on general reasoning tasks. \cite{chen2025advancing} also demonstrates that VL reasoners benefit from exposure to unimodal (text) reasoning data. Other work \citep{chen2025g1} shows that perception-enhanced reasoning chains improve the ability of VL models to play games such as Shisen-Sho. However, all these potential data interventions use different training paradigms (e.g., instruction-tuning, RL), model families and sizes, and underlying evaluations. This makes it difficult to anticipate which strategies will lead to reliable improvements in VL reasoning. To address this, we perform a comprehensive set of experiments in which several datasets are compared against each other using a fixed training algorithm, student models and their sizes, and a battery of VL reasoning evaluation datasets. Our findings help in creating the best VL reasoning dataset, \name{}, and show that model performance improves with scale, along with efficiency benefits for test-time scaling.
\section{Conclusion}

In this report, we present \name{}, a high-quality, large-scale VL reasoning dataset with chain-of-thoughts (CoTs). Future work can assess the impact of our insights on general-purpose data curation for VL training, particularly for skills that go beyond reasoning, such as VQA. Moreover, we have focused only on data curation for single images, but it would be pertinent to extend this approach to reasoning over multiple images. Overall, our work lays a strong foundation for data research in VL reasoning.

\clearpage
\newpage
\bibliographystyle{assets/plainnat}
\bibliography{paper}

\clearpage
\newpage

\beginappendix

\section{Additional Context Dataset Details and Results}
\label{app:additional_context_dataset}

We present additional details about the context datasets in Table \ref{apptab:more_details_contexts}. Specifically, we report the total number of examples originally present in each dataset. Subsequently, we perform decontamination by removing instances in which the image exactly matches any image in the validation dataset, as determined by the phash algorithm. For the context curation experiments, we randomly sample at most $50$K examples from each dataset, as shown in the third column. Furthermore, we create a filtered version of these datasets based on final answer correctness, with the resulting sizes reported in the last column.

\begin{table}[h]
\centering
\begin{tabular}{lcccc}
\toprule
  \textbf{Context Sources} & \makecell{\textbf{Total Examples}\\(in K)}  & \textbf{\# Contaminated} & \makecell{\textbf{\# Context Curation}\\(Original, in K)} & \makecell{\textbf{\# Context Curation} \\(Filtered, in K)} \\\toprule
\virl{}& 38.9  & 80& 38.9& 23.6 \\
\mathllava{}& 338.7 & 856  & 50.0& 29.0 \\
\ronevision{}& 154.6 & 508  & 50.0& 22.5 \\
\thinklite{}& 11.0  & 129  & 11.0& 4.0  \\
\llavacot{}& 98.6  & 336  & 50.0& 27.4 \\
\mmk{} & 17.6  & 1 & 17.6& 9.0\\\bottomrule 
\end{tabular}
\caption{\small{\textbf{Context source dataset statistics.} We present the total number of examples, the number of instances found to be contaminated, and the number of instances originally used in context curation experiments compared to their filtered versions.}}
\label{apptab:more_details_contexts}
\end{table}

Next, we train PLM-3B and PLM-8B on both the original and filtered versions of the source datasets, where CoTs are generated by our generator model (Llama-4-Scout). We evaluate the trained models on five downstream VL tasks and present their average accuracy in Table \ref{apptab:filtering_or_no_filtering}. We find that \virl{} and \thinklite{} benefit from final answer correctness filtering, while the original (unfiltered) datasets produce better reasoners. The best-performing version of each context source dataset is presented in the main Table \ref{tab:ranked_context_sources}.

\begin{table}[h]
\centering
\begin{tabular}{lccc}
\toprule
  \textbf{Context Sources} &  &\textbf{Original} & \textbf{Filtering} \\\toprule
\multirow{3}{*}{\virl{}}& 3B  & 38.1& 38.8\\
 & 8B  & 40.1& 41.3\\
 \rowcolor{blue!20}
\cellcolor{white} & Avg. & 39.1& \textbf{40.1}\\\hline
\multirow{3}{*}{\mathllava{}}& 3B  & 36.3& 36.2\\
 & 8B  & 39.2& 39.1\\
\rowcolor{blue!20}
\cellcolor{white} & Avg. & \textbf{37.7}& 37.6\\\hline
\multirow{3}{*}{\ronevision{}}& 3B  & 35.7& 34.8\\
 & 8B  & 38.8& 37.5\\
\rowcolor{blue!20}
\cellcolor{white} & Avg. & \textbf{37.3}& 36.2\\\hline
\multirow{3}{*}{\thinklite{}}& 3B  & 34.7& 34.6\\
 & 8B  & 38.0& 39.5\\
\rowcolor{blue!20}
\cellcolor{white} & Avg. & 36.4& \textbf{37.1}\\\hline
\multirow{3}{*}{\llavacot{}}& 3B  & 34.6& 33.8\\
 & 8B  & 37.9& 37.6\\
\rowcolor{blue!20}
\cellcolor{white} & Avg. & \textbf{36.3}& 35.7\\\hline
\multirow{3}{*}{\mmk{}} & 3B  & 34.6& 35.6\\
 & 8B  & 37.3& 36.0\\
\rowcolor{blue!20}
\cellcolor{white} & Avg. & \textbf{36.0}& 35.8 \\\bottomrule 
\end{tabular}
\caption{\small{\textbf{Impact of filtering on final answer correctness.} We present the results of training PLM-3B and PLM-8B on both the original and filtered versions of the context source datasets. Specifically, we report the average accuracy across the five evaluation datasets and bold the best-performing setup.}}
\label{apptab:filtering_or_no_filtering}
\end{table}

\section{Additional Data Intervention Details and Results}
\label{app:data_interventions}

Here, we present additional implementation details regarding the diverse data interventions. Where relevant, we also highlight experimental results across various configurations.

\paragraph{\textbf{Visual Perturbation.}} In this intervention, we apply one of three transformations {\textit{rotation}, \textit{distractor concatenation}, \textit{dominance-preserving mixup}\} to every image in the dataset. In \textit{rotation}, we rotate the image by any angle (in degrees) in this list \texttt{[0, 15, 30, 45, 60, 75, 90, 105, 120, 135, 150, 165, 180, 195, 210, 225, 240, 255, 270, 285, 300, 315, 330, 345]}. In \textit{distractor concatenation}, we treat any image from the other dataset as a distractor, and concatenate it with the original image in the \texttt{width} dimension to form a new image. In \textit{dominance-preserving mixup}, we form a new image in the input space by mixing the original image with a distractor image using \texttt{mixup image = $\alpha\times$original image + $(1-\alpha)\times$distractor image} where $\alpha \sim \mathcal{U}[0.8, 1.0)$. 

We test two approaches for applying visual perturbation to the original data: (a) \textit{replacement}, where a $50\%$ of the original data is replaced with its perturbed version, and (b) \textit{augmentation}, where the perturbed version of the original data is concatenated with the original data itself. 
For (b), we compare the performance of training VLMs on $50\%$ of the original data versus training them on the combination of this data and its visually perturbed version, effectively doubling the amount of training data at no additional cost. The results are shown in Table \ref{app:augment_curation_methods}. We find that augmenting the original data with its visually perturbed version leads to worse reasoning performance, which may be attributed to increased redundancy in the training data and the noise introduced by visual perturbations.

\begin{table}[h]
\centering
\resizebox{\textwidth}{!}{%
\begin{tabular}{cccccccc}
\toprule  
\textbf{Augmentation Expts.} &  \textbf{Models}& \textbf{Average} & \textbf{MathVerse} & \textbf{MathVista} & \textbf{MathVision} & \textbf{MMMU-Pro} & \textbf{We-Math} \\\toprule
\multirow{3}{*}{50\% Original Data}  & 3B  & 35.1 & 28.7	& 52.7	& 23.0	& 25.0& 	46.1\\
& 8B  &  40.0 & 32.6& 61.5	&24.3&	28.4	&53.2\\
& \cellcolor{blue!20}Avg. & \cellcolor{blue!20}\textbf{37.6} &\cellcolor{blue!20} 30.6	& \cellcolor{blue!20}57.1	& \cellcolor{blue!20}23.7	& \cellcolor{blue!20}26.7	&  \cellcolor{blue!20}49.7 \\\hline
\multirow{3}{*}{\makecell{50\% Original Data \\+ 50\% Visual Perturbation}}& 3B  & 34.6 & 28.3 & 52.6 & 19.1	& 24.5	& 48.5 \\
& 8B  & 39.2 & 31.9	&62.0	&23.4	&27.4	&51.4 \\
& \cellcolor{blue!20}Avg. & \cellcolor{blue!20}36.9 &\cellcolor{blue!20}30.1&	\cellcolor{blue!20}57.3	&\cellcolor{blue!20}21.2	&\cellcolor{blue!20}26.0&	\cellcolor{blue!20}50.0\\
\hline
\multirow{3}{*}{\makecell{50\% Original Data \\+ 50\% Text-Rich Images}} & 3B  & 34.8 &28.7&	55.4	&16.1&	25.4	&48.2  \\
& 8B  & 39.4 & 32.9	&59.5	&22.7	&28.1	&53.7 \\
& \cellcolor{blue!20}Avg. & \cellcolor{blue!20}37.1 &\cellcolor{blue!20}30.8	&\cellcolor{blue!20}57.5&	\cellcolor{blue!20}19.4&	\cellcolor{blue!20}26.8	&\cellcolor{blue!20}51.0  \\
\bottomrule
\end{tabular}%
}
\caption{\small{\textbf{Data intervention augmentation results.} }}
\label{app:augment_curation_methods}
\end{table}

\paragraph{\textbf{Text-Rich Images.}} In this intervention, we render the original image and question from the dataset into a new in the raw space on randomly sampled background colors, text fonts and colors. We use functions from the \texttt{PIL} python library to achieve this task. In particular, we provide our rough implementation in Listing \ref{lst:text_rich_code}. Similar to visual perturbations, the \textit{text-rich images} intervention can be used in two ways: (a) \textit{replacement} on $50\%$ of original data, and (b) \textit{augmentation}. We highlight the results for \textit{augmentation} strategy in Table \ref{app:augment_curation_methods}. In particular, we find that augmenting the data with text-rich images does not improve the reasoning performance over the half the original data (row 1 vs row 3).

\paragraph{\textbf{Perceptual Redundancy.}} This strategy filters the instances in the original data if the blind generator model can solve the question correctly i.e., without access to the image. We observe that this strategy removes $42\%$ of the instances in the data curation experiments.

\paragraph{\textbf{Shallow Perception.}} This strategy is a stricter version of the perceptual redundancy where the instances that can be solved correctly using a blind generator model with access to image captions are filtered. In our experiments, we observe that this strategy removes $55\%$ of the data.

\begin{table}[h]
\centering
\resizebox{\textwidth}{!}{%
\begin{tabular}{lccccccc}
\toprule  
\textbf{Caption and Solve Variants} &  \textbf{Models}& \textbf{Average} & \textbf{MathVerse} & \textbf{MathVista} & \textbf{MathVision} & \textbf{MMMU-Pro} & \textbf{We-Math} \\\toprule
\multirow{3}{*}{\begin{tabular}[c]{@{}l@{}}I $\rightarrow$ C\\ (I, Q) $\rightarrow$ S\end{tabular}} & 3B  &  39.7& 35.7  & 56.6  & 24.7& 26.7 & 55.0 \\
  & 8B  &  43.0& 38.3  & 63.2  & 26.3& 30.4 & 56.9 \\
  & \cellcolor{blue!20}Avg. &\cellcolor{blue!20} \textbf{41.4}&\cellcolor{blue!20} 37.0  &\cellcolor{blue!20} 59.9  & \cellcolor{blue!20}25.5& \cellcolor{blue!20}28.6 & \cellcolor{blue!20}56.0 \\\hline
\multirow{3}{*}{(I, Q) $\rightarrow$ (C, S)}  & 3B  & 37.8& 32.5  & 55.1  & 22.4& 25.8 & 53.4\\
  & 8B  & 42.3& 38.4  & 60.4  & 27.6& 29.0 & 55.8 \\
  & \cellcolor{blue!20}Avg. & \cellcolor{blue!20}40.0& \cellcolor{blue!20}35.5  & \cellcolor{blue!20}57.7  & \cellcolor{blue!20}25.0& \cellcolor{blue!20}27.4 &\cellcolor{blue!20} 54.6\\\hline
\multirow{3}{*}{\begin{tabular}[c]{@{}l@{}}I $\rightarrow$ C\\ (I, Q, C) $\rightarrow$ S\end{tabular}} & 3B  & 36.6& 31.2  & 52.3  & 23.1& 24.4 & 52.1 \\
  & 8B  & 38.9& 30.2  & 60.5  & 22.0& 29.4 & 52.6 \\
  & \cellcolor{blue!20}Avg. & \cellcolor{blue!20}37.8& \cellcolor{blue!20}30.7  & \cellcolor{blue!20}56.4  & \cellcolor{blue!20}22.5& \cellcolor{blue!20}26.9 & \cellcolor{blue!20}52.3 \\\bottomrule
\end{tabular}%
}
\caption{\small{\textbf{Caption and Solve variants.} We present the accuracy of models trained with different data generation variants for the \textit{caption and solve} intervention. Here, I $\rightarrow$ C means that the teacher model takes image I as input and outputs the caption C. Furthermore, (I, Q) $\rightarrow$ S indicates that the teacher model takes the (image, question) pair as input and outputs the solution S. We note that the student VL reasoner model is always trained to generate the caption C and solution S jointly, conditioned on the image and question pair: (I, Q) $\rightarrow$ (C, S).}}
\label{app:caption_solve_exploration}
\end{table}

\paragraph{\textbf{Caption and Solve.}} In this strategy, the chain-of-thought (CoT) is augmented to first generate an image caption, providing auxiliary perceptual signals before addressing the given question. Specifically, we explore three approaches for creating such data using our generator model: (a) synthesizing the image caption and problem-solving steps independently i.e., $I_j^{cap} = \mathcal{G}(I_j)$ and $C_j = \mathcal{G}(I_j, Q_j)$, (b) generating the image caption first, followed by problem-solving  $(I_j^{cap}, C_j) = \mathcal{G}(I_j, Q_j)$, and (c) generating the image caption and then providing it as additional context for problem-solving i.e., $C_j = \mathcal{G}(I_j, Q_j, I_j^{cap})$. Ultimately, the VLMs are trained in an identical fashion across all three strategies; that is, the input consists of (image, question), and the CoT comprises (caption, solution). We present the results from these three strategies in Table \ref{app:caption_solve_exploration}. We observe that the first strategy produces CoTs that train the strongest VL reasoners. This approach is also the most efficient in terms of scalable data generation, as the caption and solution can be generated in parallel and later merged for training VL reasoners. 

\paragraph{\textbf{Text-Only Reasoning.}} In this strategy, we augment the original VL reasoning data with the text-only reasoning data. In particular, we choose a state-of-the-art reasoning data, \ot{} \cite{guha2025openthoughts}. The CoTs in the original dataset are collected the QwQ-32B reasoning model.\footnote{\url{https://qwenlm.github.io/blog/qwq-32b/}} To ensure consistency between the generator models, we re-annotate the entire 1.04 million samples with Llama-4-Scout.

\paragraph{\textbf{Increased Distractors.}} In this strategy, we re-write the dataset to increase the number of distractors (Listing \ref{lst:llama4_distractor}). In our experiments, we consider the \textit{replacement} strategy with this intervention on $50\%$ of original data.

\paragraph{\textbf{Length.}} In this strategy, we bias our dataset to elicit long CoTs. Specifically, we compute the median CoT length and divide the dataset into two equal halves i.e., instances with CoTs lesser than the median and the ones more than the median. 

\paragraph{\textbf{Difficulty.}} In this strategy, we classify each instance in the dataset according to its difficulty level. In our dataset of size $50$K, the distribution of difficulty levels is \texttt{\{Level 2: 15.1K, Level 1.5: 14.4K, Level 1: 11.8K, Level 4: 3.6K, Level 3: 3.5K, Level 4+: 0.5K\}}. To increase the impact of harder examples, we select $3500$ samples from each difficulty level and train on this uniform-difficulty dataset.

\section{Self-Improvement}
\label{app:self_improvement_experiment}

Here, we aim to study the impact of training the teacher model, Llama-4-Scout ($109A17$B MoE), on the self-generated reasoning data, \name{}. Specifically, we take $50$K subset of this data and finetune this model using low-rank adaptation (LoRA) with the \texttt{Llama-factory} library.\footnote{\url{https://github.com/hiyouga/LLaMA-Factory/pull/7611}} We present the results across downstream evaluation datasets in Table \ref{tab:Llama-4}. We find that parameter-efficient finetuning of Llama-4 on the \name{} dataset leads to a relative gain of $3.7$pp over the base model. Importantly, we observe improvements across most of the evaluation datasets indicating the strong self-improvement capabilities. Future work can focus on training much stronger models on using our high-quality and large-scale dataset.

\begin{table}[t]
\centering
\begin{tabular}{lcccccccc}
\toprule
& \rotatebox{90}{\textbf{Average}} 
& \rotatebox{90}{\makecell{\textbf{MathVerse}}} 
& \rotatebox{90}{\makecell{\textbf{MathVista}}} 
& \rotatebox{90}{\makecell{\textbf{MathVision}}} 
& \rotatebox{90}{\makecell{\textbf{MMMU-Pro}}} 
& \rotatebox{90}{\makecell{\textbf{We-Math}}} 
& \rotatebox{90}{\textbf{MATH500}} 
& \rotatebox{90}{\makecell{\textbf{GPQA}}} \\
\toprule  
Llama-4-Scout & 59.3& 53.9  & 69.3  & 32.6& 51.2 & 72.9& 81.2& 54.0 \\
\rowcolor{blue!20}Llama-4-Scout (\name{}) & \textbf{61.5}& \textbf{54.3} & \textbf{70.9}  & \textbf{37.5}& \textbf{51.7} & 72.8& \textbf{83.0}& \textbf{60.6}\\
\bottomrule
\end{tabular}
\caption{\small{\textbf{Self-improvement results.} We train the generator model, Llama-4-Scout, on a subset of the \name{} dataset and report its accuracy on diverse evaluation datasets.}}
\label{tab:Llama-4}
\end{table}

\section{Additional Training and Evaluation Setup}
\label{app:training_setup}

\paragraph{\textbf{PLM setup.}} In this work, we predominantly train Perception Language Models (PLMs) \cite{cho2025perceptionlm} for our data curation and large-scale experiments. Specifically, the PLMs use a Llama-3.2 LLM backbone that processes visual input via a perception encoder \cite{bolya2025perception}. We train PLMs using their publicly available codebase, starting with the stage-3 configurations.\footnote{\url{https://github.com/facebookresearch/perception_models}} In particular, we perform full finetuning of all parameters of the PLMs including the vision encoder and the LLM backbone. Due to the sheer number of experiments, we use the default peak learning rate (LR) of $4\mathrm{e}{-5}$ for PLM-1B and PLM-3B, and $1\mathrm{e}{-5}$ for PLM-8B. The warmup ratio is set to $0.1$ followed by cosine decay upto $10\%$ of the peak learning rate. Furthermore, we train our models for $5$ epochs across all experiments. In our data curation experiments, we set a global batch size of $2048$ for PLM-1B and PLM-3B, and $1024$ for PLM-8B, while we double the global batch sizes for the large-scale runs. All experiments are performed on multiple nodes i.e., either $8$ or $16$ A100 (80GB) nodes.

\paragraph{\textbf{RL Training.}} In Table~\ref{tab:rl_main_results}, we present a round of RL training using the GRPO algorithm. Specifically, we use the default implementation from the TRL library.\footnote{\url{https://github.com/huggingface/trl/blob/main/examples/scripts/grpo_vlm.py}} Since this work is focused on supervised finetuning, we conduct lightweight GRPO training (1 epoch) to demonstrate its effectiveness as a cold start for RL. In particular, we use a verifiable multimodal dataset.\footnote{\url{lmms-lab/multimodal-open-r1-8k-verified}} We set the number of generations per prompt in the GRPO algorithm to four. Further, we use the peak learning rate of $1\mathrm{e}{-6}$ works well for PLM-\name{} SFT models in the TRL configuration.

\paragraph{\textbf{Evaluation.}} To ensure consistent prompting across all benchmarks, we query our model with \texttt{`Your final answer MUST BE put in \boxed{boxed}.'} Furthermore, we score our models using the rule-based verifier \texttt{mathruler} for scalability and fast iteration following \cite{deng2025openvlthinker}.\footnote{\url{https://github.com/hiyouga/MathRuler}, \url{https://github.com/yihedeng9/OpenVLThinker}} We notice that the original \mverse{} free-form questions often contain numerics followed by units (e.g., celsius, cm), which makes rule-based verification more challenging. To address this, we use an LLM to extract only the final answer values and consider them as the ground-truth answers. 

\begin{figure}[h]
\centering
 \begin{subfigure}[h]{0.3\textwidth}
\centering
\includegraphics[width=\textwidth]{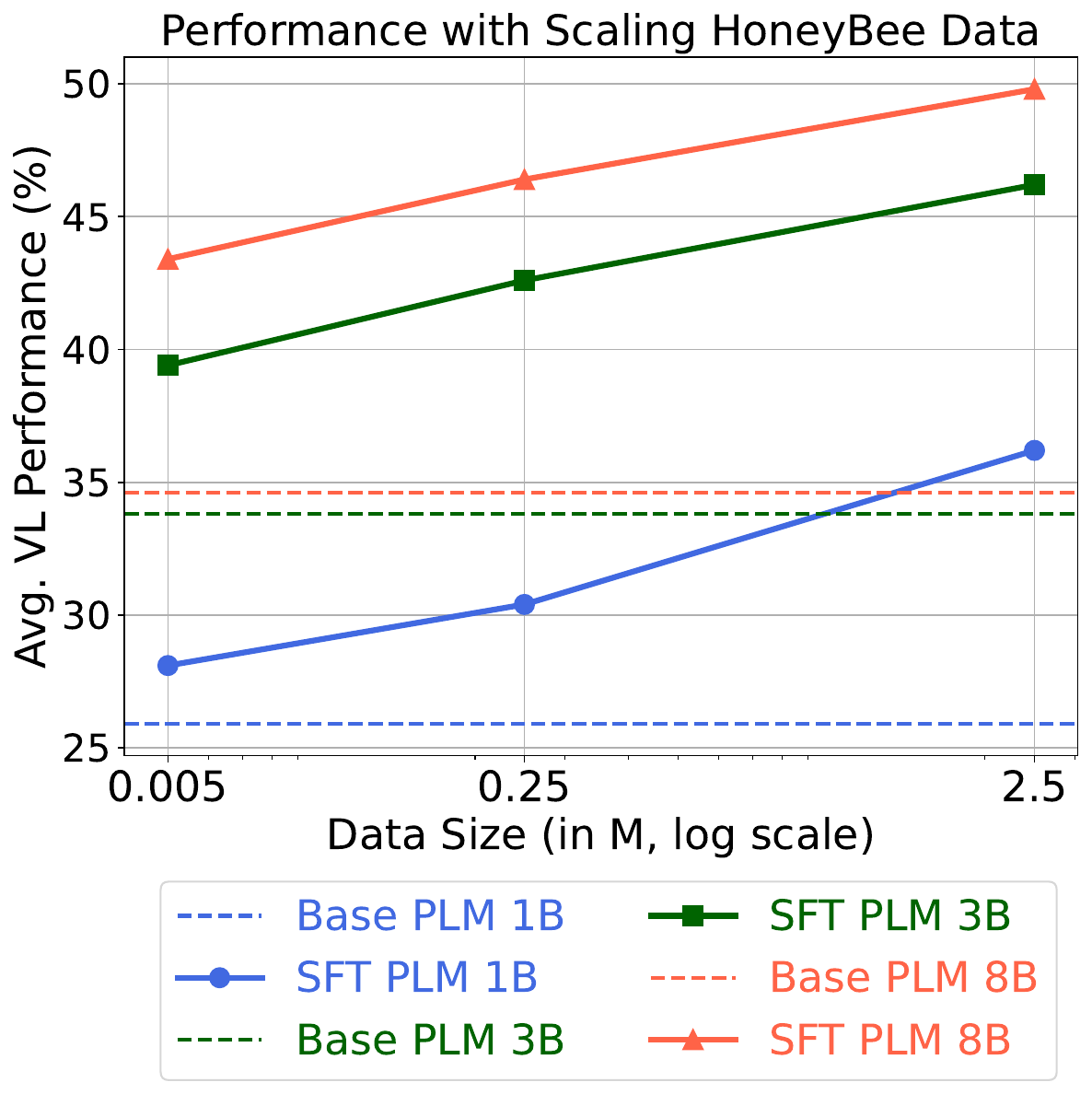}
\caption{\small{}}
\label{fig:}
 \end{subfigure}
 \begin{subfigure}[h]{0.32\textwidth}
\centering
\includegraphics[width=\textwidth]{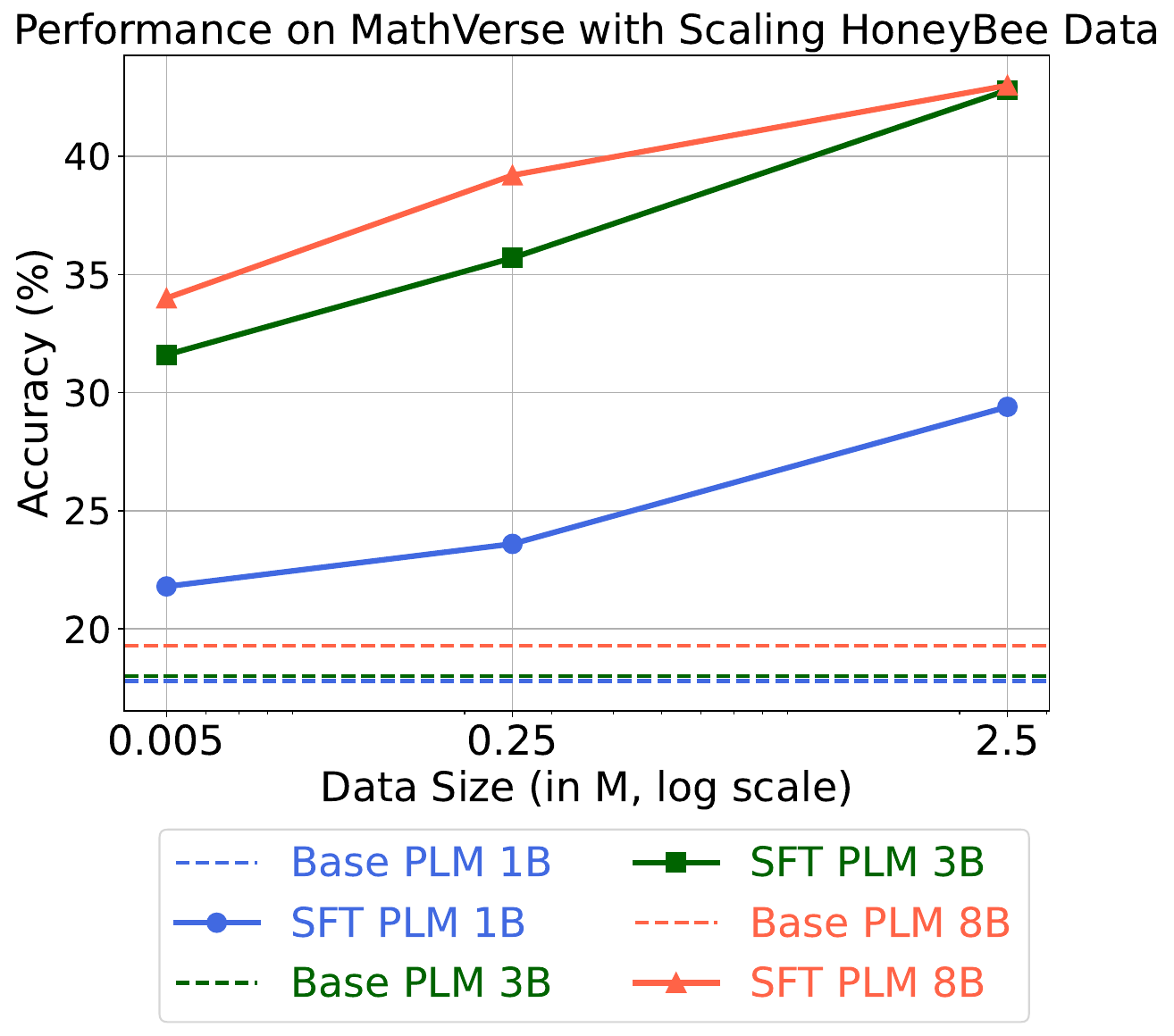}
\caption{\small{}}
\label{fig:}
 \end{subfigure}
 \begin{subfigure}[h]{0.32\textwidth}
 \centering
\includegraphics[width=\textwidth]{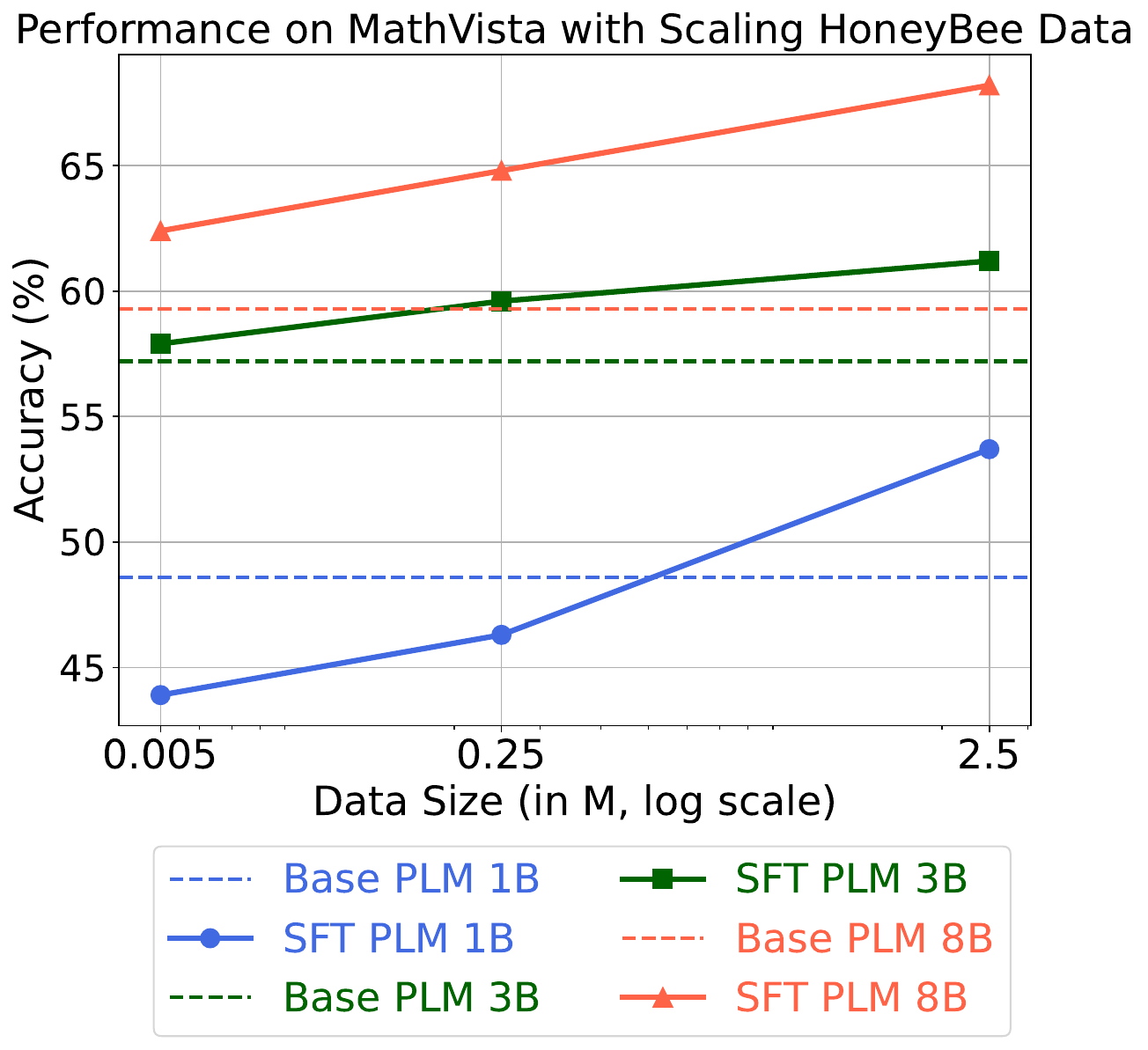}
\caption{\small{}}
 \label{fig:}
 \end{subfigure}
 \begin{subfigure}[h]{0.32\textwidth}
\centering
\includegraphics[width=\textwidth]{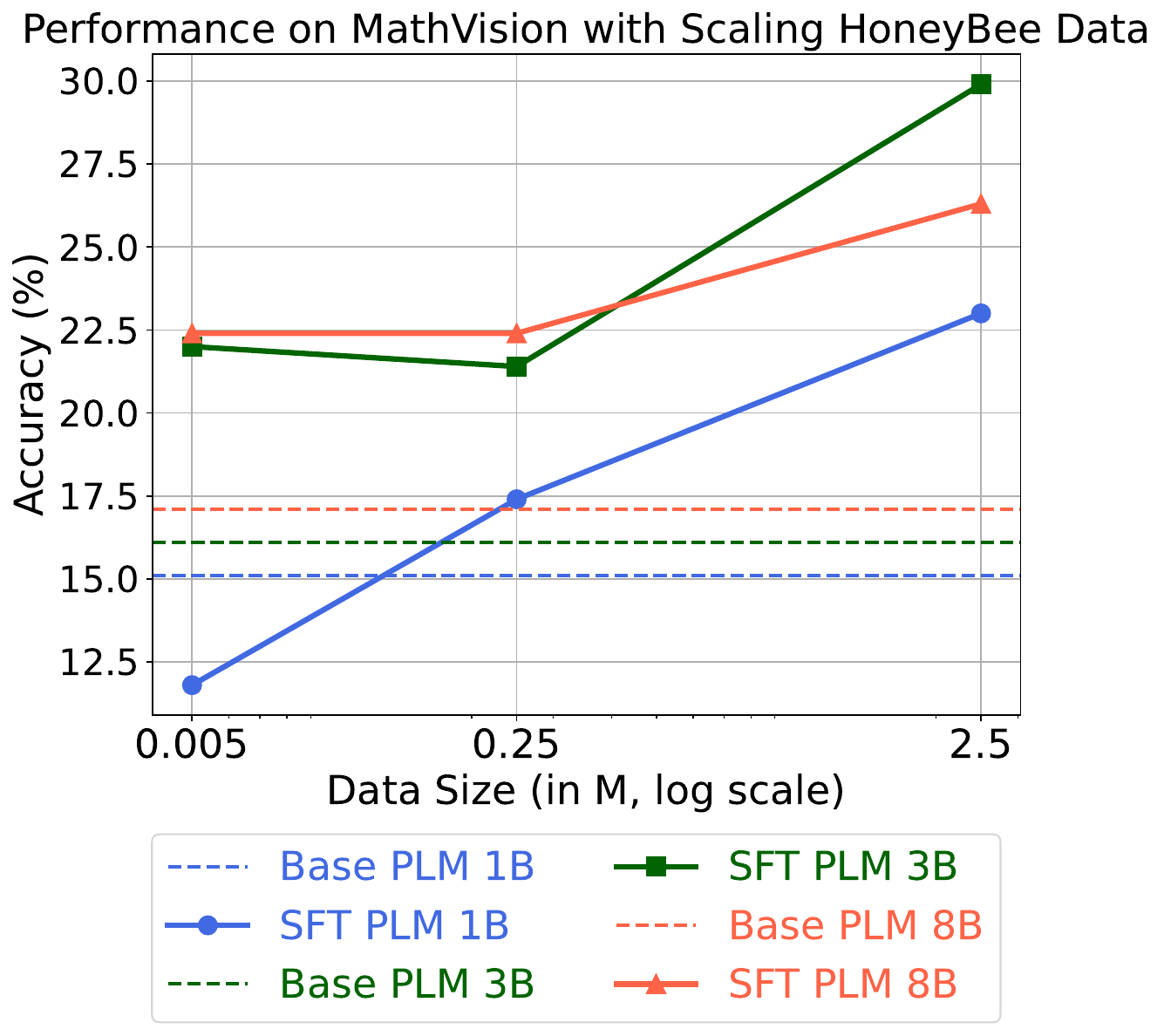}
\caption{\small{}}
\label{fig:}
 \end{subfigure}
\quad
 \begin{subfigure}[h]{0.32\textwidth}
 \centering
\includegraphics[width=\textwidth]{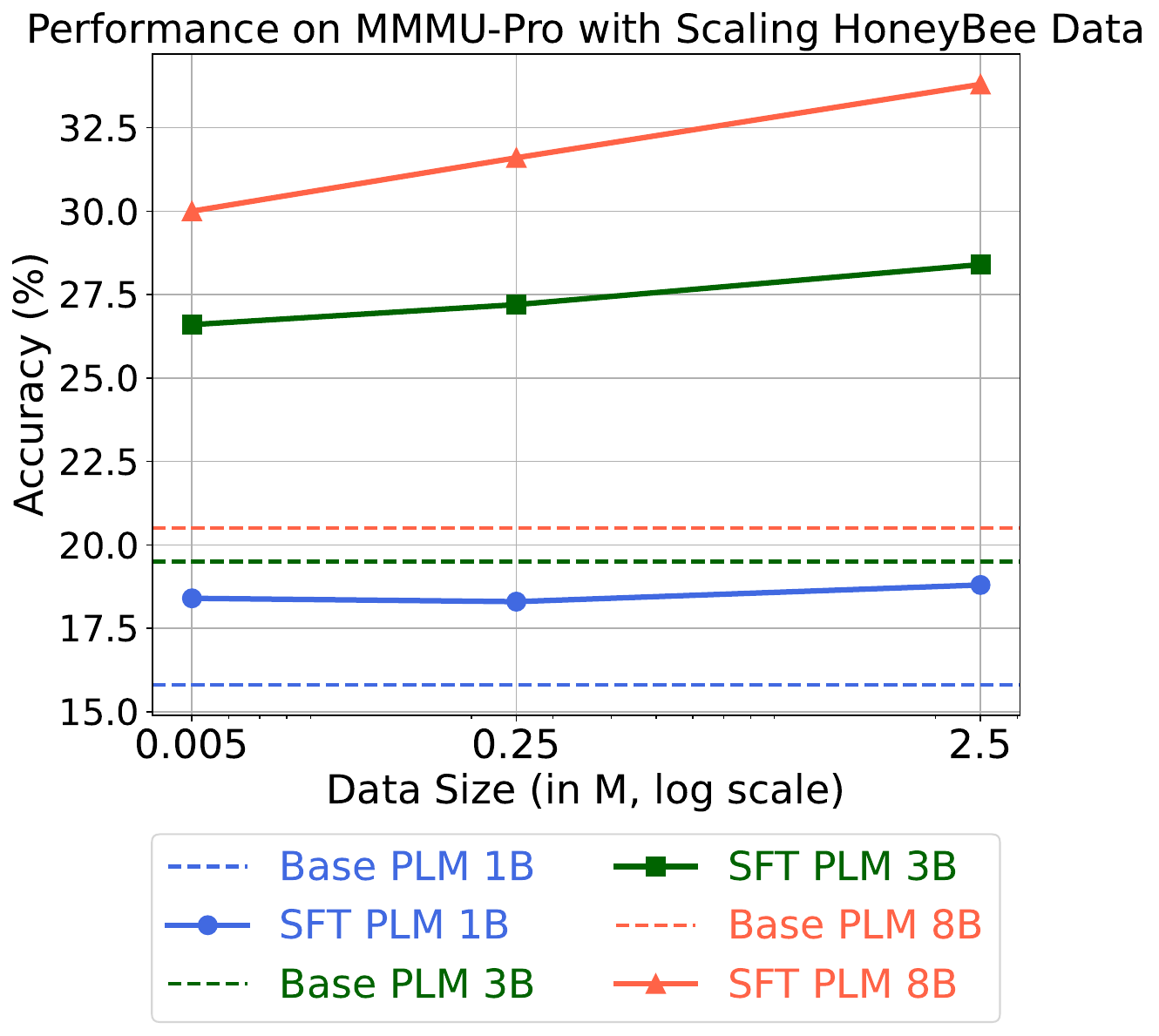}
\caption{\small{}}
 \label{fig:}
 \end{subfigure}
 \begin{subfigure}[h]{0.32\textwidth}
 \centering
 \includegraphics[width=\textwidth]{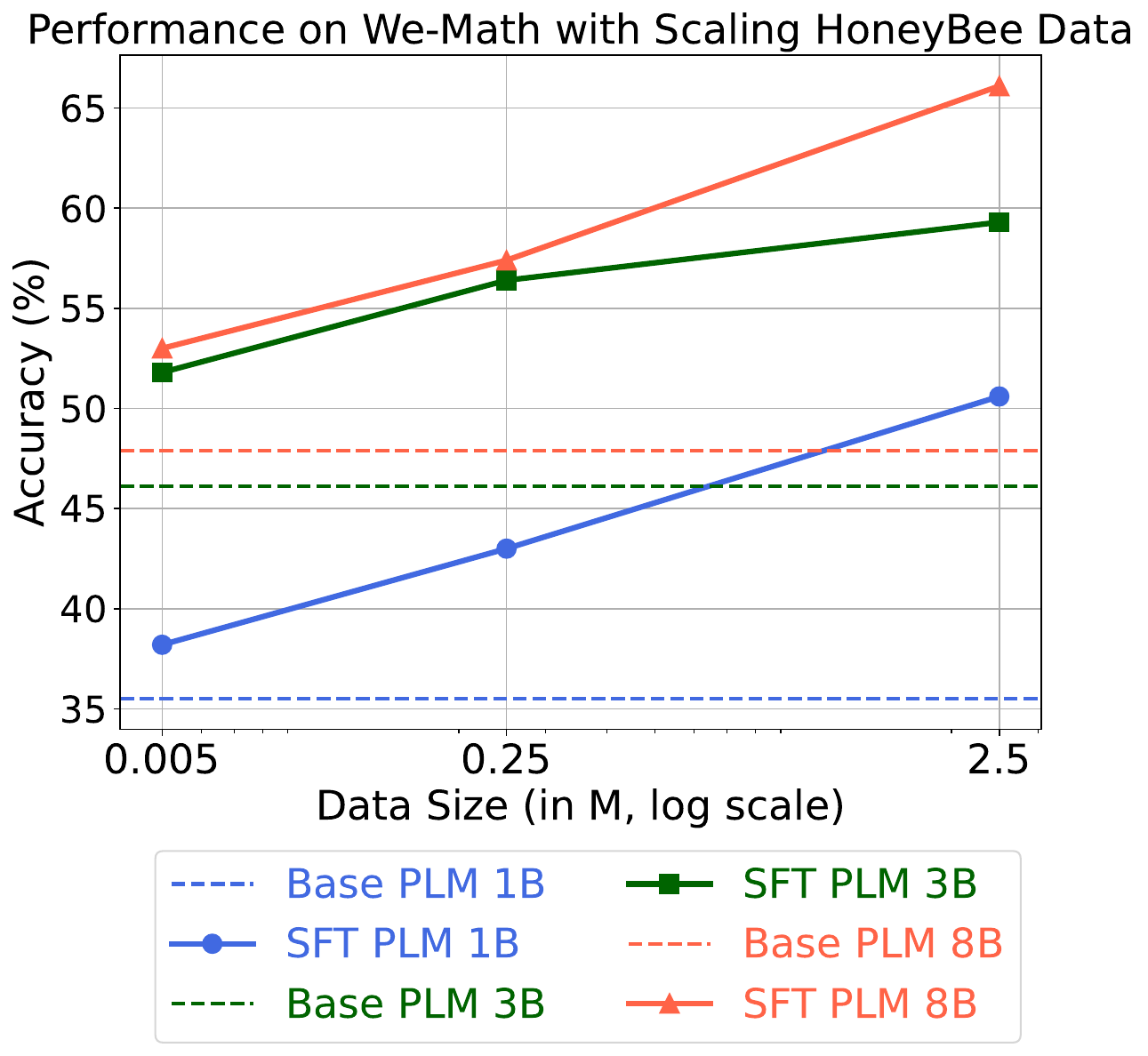}
 \caption{\small{}}
 \label{fig:}
 \end{subfigure}
\quad
 \begin{subfigure}[h]{0.32\textwidth}
 \centering
  \includegraphics[width=\textwidth]{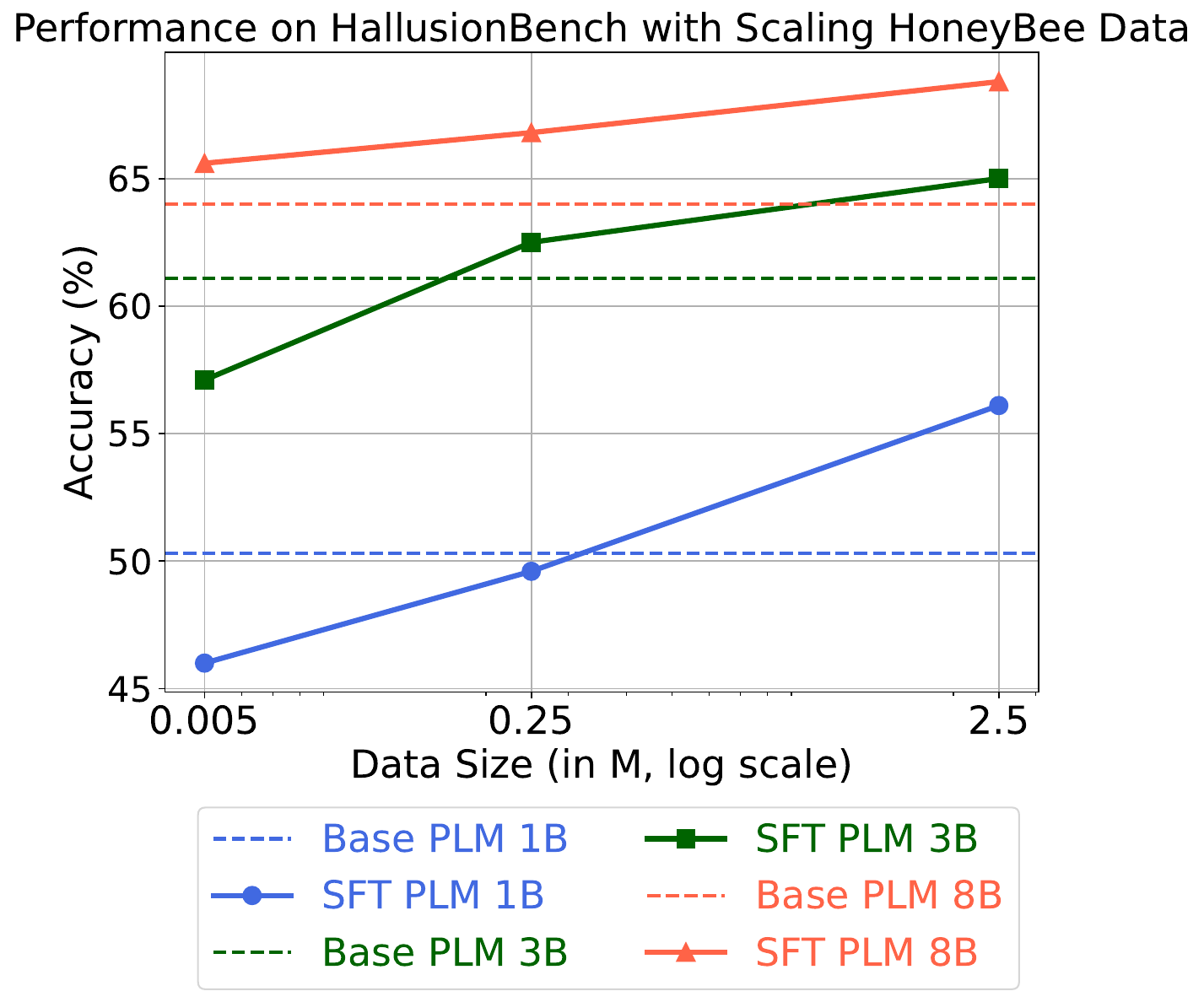}
  \caption{\small{}}
 \label{fig:}
 \end{subfigure}
 \begin{subfigure}[h]{0.32\textwidth}
 \centering
 \includegraphics[width=\textwidth]{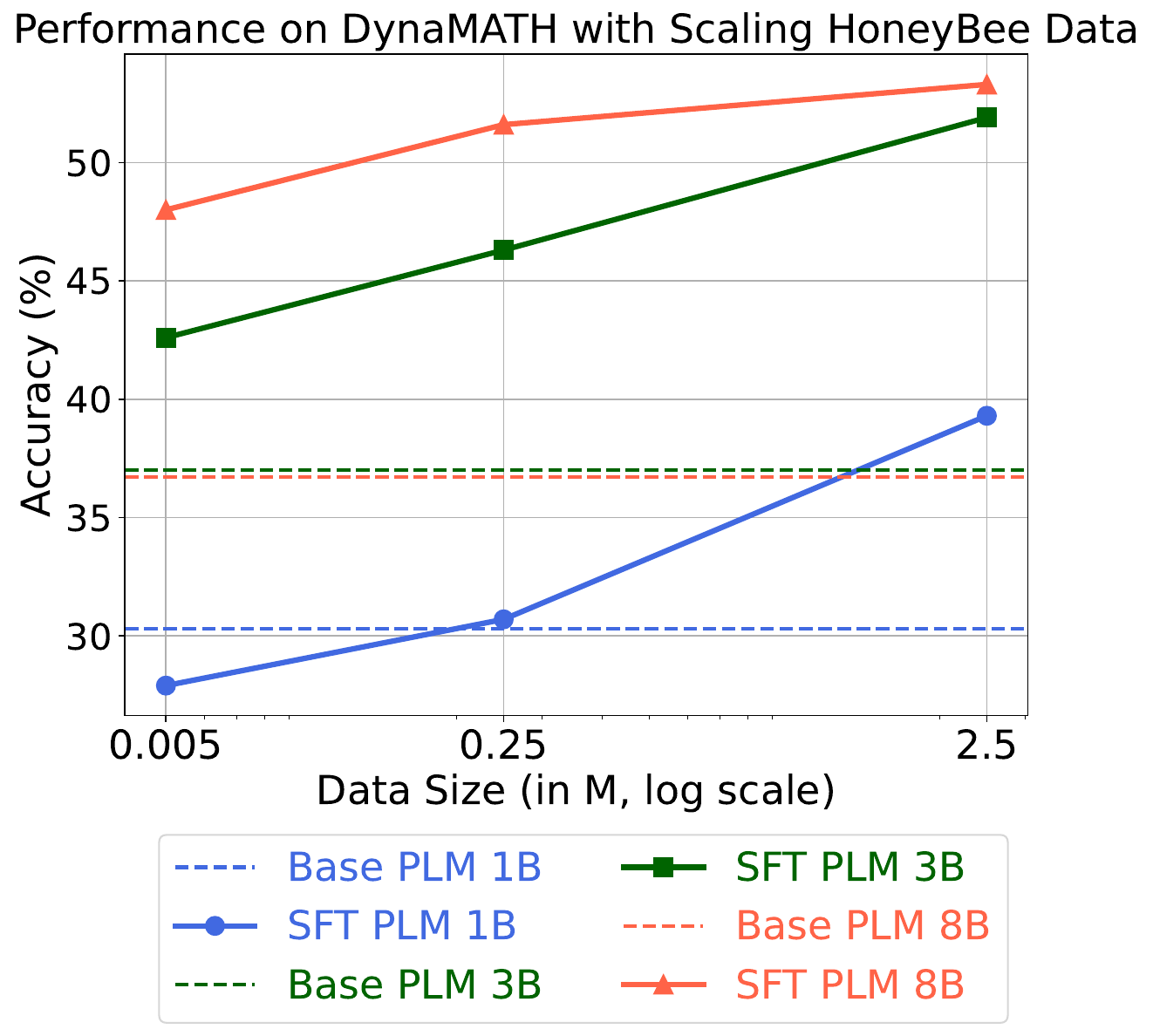}
 \caption{\small{}}
 \label{fig:}
 \end{subfigure}
 \begin{subfigure}[h]{0.32\textwidth}
 \centering
  \includegraphics[width=\textwidth]{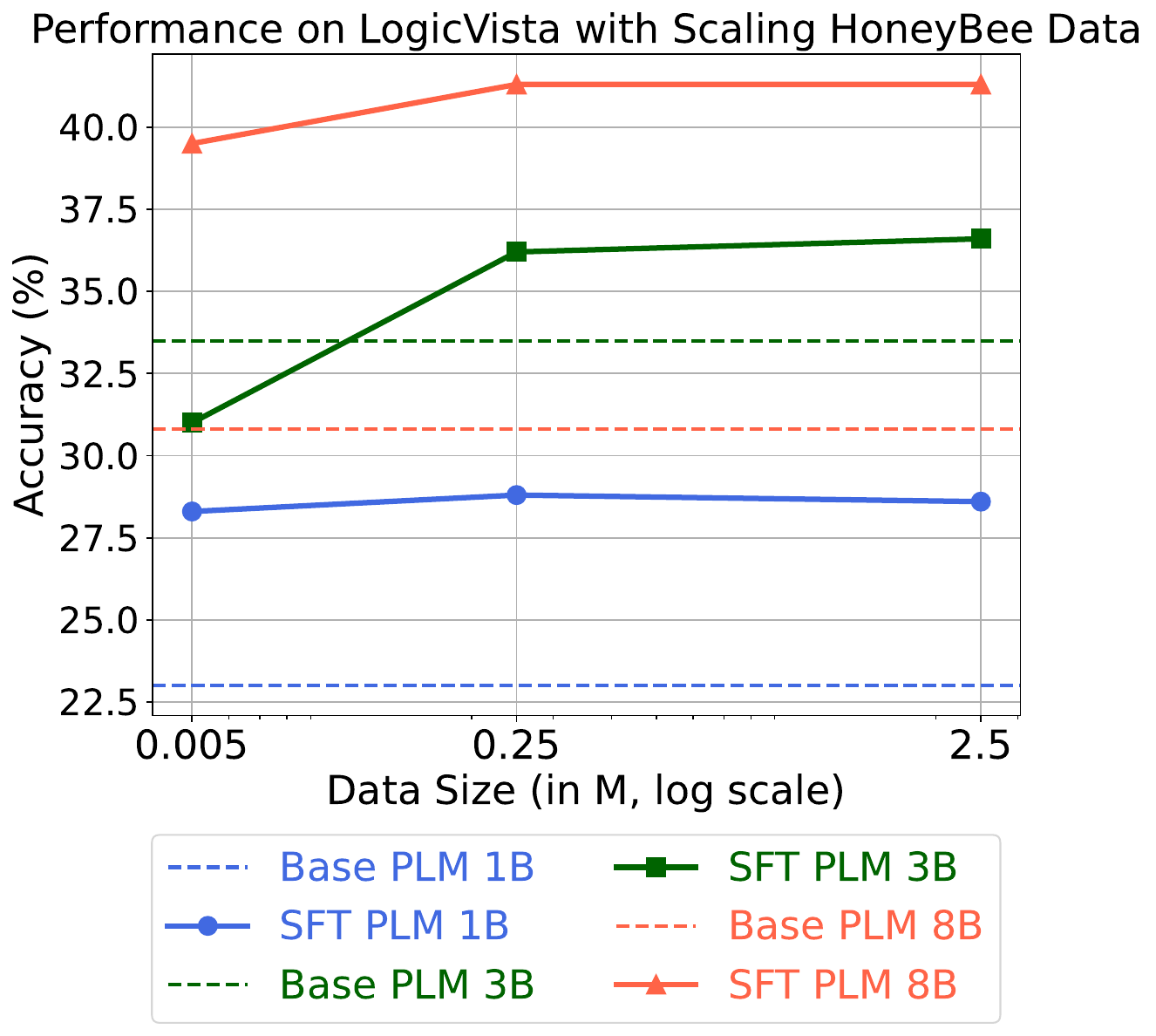}
  \caption{\small{}}
 \label{fig:}
 \end{subfigure}
 \begin{subfigure}[h]{0.32\textwidth}
 \centering
 \includegraphics[width=\textwidth]{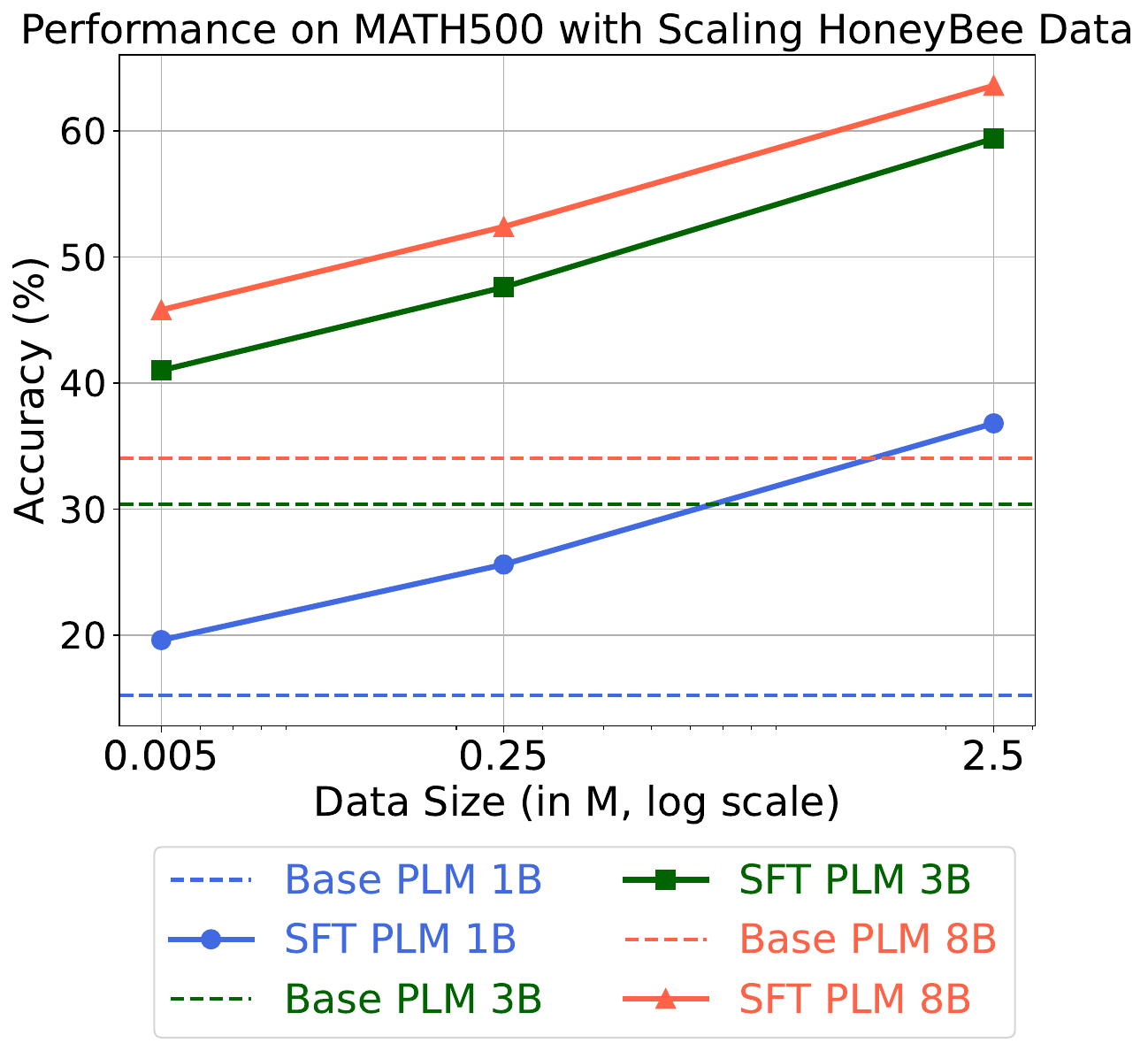}
 \caption{\small{}}
 \label{fig:}
 \end{subfigure}
 \begin{subfigure}[h]{0.32\textwidth}
 \centering
\includegraphics[width=\textwidth]{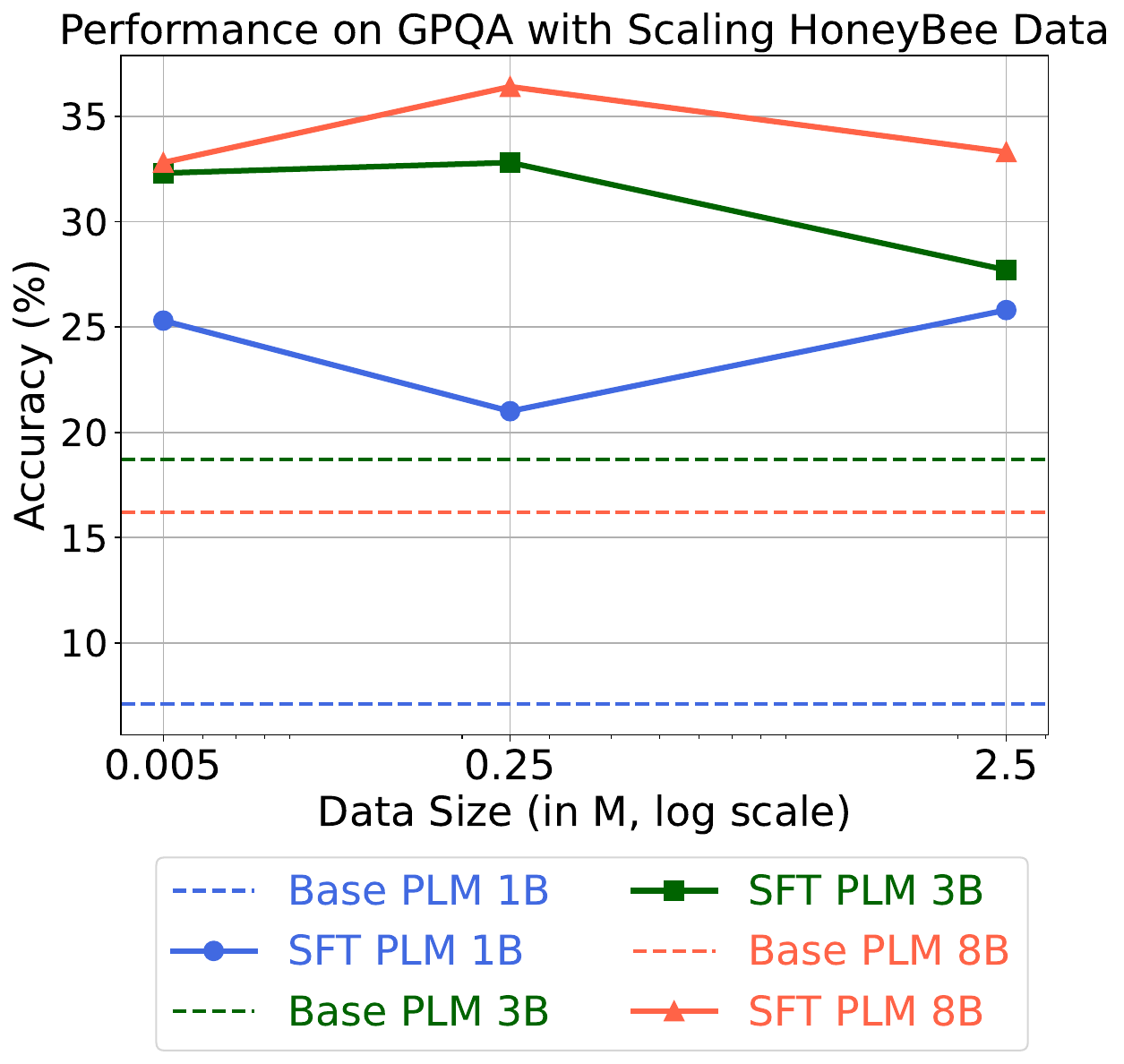}
\caption{\small{}}
 \label{fig:}
 \end{subfigure}
 \caption{\small{\textbf{Scaling trends across individual evaluation datasets.} We train PLMs ($1$B-$8$B) on diverse subsets of the \name{} data and evaluate them on individual datasets. The first plot (first row, first column) is the average performance across the ten evaluation datasets.}}
 \label{fig:scaling_per_dataset}
\end{figure}

\section{Prompts}
\label{app:prompts}

\begin{tcolorbox}[
breakable,
width=\textwidth,
colback=white,
colframe=blue!50,
title={\centering \textbf{CoT generation prompt for a given (image, question) pair.}}
]
\footnotesize
\begin{lstlisting}[
breaklines=true,
postbreak={},
breakindent=0pt,
label={lst:llamagen},
frame=none,
numbers=left,  % Enable line numbers
numbersep=15pt,% Space between numbers and code
xleftmargin=20pt,  % Left margin for code alignment
stepnumber=1% Show every line number
]
Think about the reasoning process required to solve the question in detail based on the given image and then provide the answer. The reasoning process should be enclosed within <think> and </think> tags followed by the summary of the answer.

In your reasoning, incorporate the following elements:

- Planning: Systematically outline the steps before executing them.
- Evaluation: Critically assess each intermediate step.
- Reflection: Reconsider and refine your approach as needed.
- Exploration: Consider alternative solutions.

If the question is multiple-choice, present the final answer as the option letter A, B, C, ... F.

Enclose the final answer in \\boxed{}.

Question: <Question>
\end{lstlisting}
\end{tcolorbox}

\begin{tcolorbox}[
breakable,
width=\textwidth,
colback=white,
colframe=blue!50,
title={\centering \textbf{Prompt to generate new question for a given (image, question) pair.}}
]
\footnotesize
\begin{lstlisting}[
breaklines=true,
postbreak={},
breakindent=0pt,
label={lst:llamaqgen},
frame=none,
numbers=left,  % Enable line numbers
numbersep=15pt,% Space between numbers and code
xleftmargin=20pt,  % Left margin for code alignment
stepnumber=1% Show every line number
]
Help the user create a new math problem that inspired by the original one related to the provided image. 

- Make sure the new problem is logical and can be solved. 
- The new question should be directly based on the image, not an independent question.
- Do not solve the original question.
- Conclude your response with "New Question: ". If necessary, please include options or choices in the new question.
- Do not provide the answer in your new question.

Original Question: <Question>
\end{lstlisting}
\end{tcolorbox}

\begin{tcolorbox}[
breakable,
width=\textwidth,
colback=white,
colframe=blue!50,
title={\centering \textbf{Prompt to generate a detailed caption for a given image.}}
]
\footnotesize
\begin{lstlisting}[
breaklines=true,
postbreak={},
breakindent=0pt,
label={lst:llamacaption},
frame=none,
numbers=left,  % Enable line numbers
numbersep=15pt,% Space between numbers and code
xleftmargin=20pt,  % Left margin for code alignment
stepnumber=1% Show every line number
]
Provide a detailed caption of the given image.

Caption: 
\end{lstlisting}
\end{tcolorbox}

\begin{tcolorbox}[
breakable,
width=\textwidth,
colback=white,
colframe=blue!50,
title={\centering \textbf{Prompt to rewrite the original (question, CoT) pair with distractors.}}
]
\footnotesize
\begin{lstlisting}[
breaklines=true,
postbreak={},
breakindent=0pt,
label={lst:llama4_distractor},
frame=none,
numbers=left,  % Enable line numbers
numbersep=15pt,% Space between numbers and code
xleftmargin=20pt,  % Left margin for code alignment
stepnumber=1% Show every line number
]
Please rewrite the question shown in the image to include a total of 10 options.
- Rewrite the given question to include a total of 10 options. If the original question does not have any options, create and add 10 options to it. 
- If the original question has options then the original options should be shuffled (e.g., content of option A goes to option D etc.)
- Rewrite the predicted answer in its entirety, maintaining the same step-by-step structure and content as the original answer.
- Ensure that the final answer should be in the format of \\boxed{Option Letter} , where "Option Letter" corresponds to the correct choice.
- Do not attempt to solve the question at any time, just focus on rewriting.

Question: <Question>

Answer: <Answer>

Respond with "Rewrite Question: " and "Rewrite Answer: ".
\end{lstlisting}
\end{tcolorbox}

\begin{tcolorbox}[
breakable,
width=\textwidth,
colback=white,
colframe=blue!50,
title={\centering \textbf{Code snippet to create text-rich images.}}
]
\footnotesize
\begin{lstlisting}[
breaklines=true,
postbreak={},
breakindent=0pt,
label={lst:text_rich_code},
frame=none,
numbers=left,  % Enable line numbers
numbersep=15pt,% Space between numbers and code
xleftmargin=20pt,  % Left margin for code alignment
stepnumber=1% Show every line number
]
from PIL import Image, ImageDraw, ImageFont

# Text wrapping function
def wrap_text(text, font, max_width):
    """Wrap text to fit within max_width pixels"""
    words = text.split()
    lines = []
    current_line = []
    
    for word in words:
        # Test if adding this word would exceed max width
        test_line = ' '.join(current_line + [word])
        bbox = draw.textbbox((0, 0), test_line, font=font)
        text_width = bbox[2] - bbox[0]
        
        if text_width <= max_width:
            current_line.append(word)
        else:
            if current_line:
                lines.append(' '.join(current_line))
                current_line = [word]
            else:
                # Single word is too long, force it
                lines.append(word)
    
    if current_line:
        lines.append(' '.join(current_line))
    
    return lines


def get_text_rich_image(image_path, question, save_path, background_colors, text_colors, text_fonts):

    """
    background_colors, text_colors: list of RGB codes
    text_fonts: list of .ttf files
    """
    
    ## blank canvas
    img_width = random.choice(img_widths)
    img_height = 512 * 4
    background_color = random.choice(background_colors)

    # Create a new image with the background color
    img = Image.new('RGB', (img_width, img_height), color=background_color)
    draw = ImageDraw.Draw(img)

    ## embed text
    text_color = random.choice(text_colors)
    font = ImageFont.truetype(random.choice(text_fonts), 16)  # Linux

    margin = 20
    max_text_width = img_width - (2 * margin)
    line_spacing = 5

    # Wrap the text
    embedded_question = question.replace("\n", " ")
    wrapped_lines = wrap_text(embedded_question, font, max_text_width)

    # Calculate line height
    sample_bbox = draw.textbbox((0, 0), "Sample", font=font)
    line_height = sample_bbox[3] - sample_bbox[1]

    # Draw the wrapped text
    y_position = margin
    for line in wrapped_lines:
        draw.text((margin, y_position), line, fill=text_color, font=font)
        y_position += line_height + line_spacing

    ## embed image
    overlay_image = Image.open(image_path)

    original_overlay_img_width = overlay_image.width
    if original_overlay_img_width > int(0.8 * img_width):
        overlay_width = int(0.8 * img_width)
    else:
        overlay_width = original_overlay_img_width
    original_width, original_height = overlay_image.size
    
    aspect_ratio = original_height / original_width
    overlay_height = int(overlay_width * aspect_ratio)

    overlay_img_resized = overlay_image.resize((overlay_width, overlay_height), Image.Resampling.LANCZOS)

    if overlay_img_resized.mode == 'RGBA':
        img.paste(overlay_img_resized, (int(0.1 * img_width), y_position + margin), overlay_img_resized)
    else:
        img.paste(overlay_img_resized, (int(0.1 * img_width), y_position + margin))

    height = y_position + overlay_height + (2 * margin)

    ## save image
    img = img.crop((0, 0, img_width, height))
    img.save(save_path)

\end{lstlisting}
\end{tcolorbox}

\section{Qualitative Examples}
\label{app:qualitative_examples}

We show several qualitative examples of newly generated questions for a given image in Figure \ref{fig:scale_question_example_1} and Figure \ref{fig:scale_question_example_2}. These examples highlight that the synthetic questions can be quite diverse and reasonable for training VL reasoners. Furthermore, we present a few qualitative examples from the \name{} dataset in Figure \ref{fig:qual_example_1} and Figure \ref{fig:qual_example_2}. These examples demonstrate that the CoTs in our dataset are formed by concatenating the captions and problem solutions in the raw text space. 

\begin{table*}[th!]
\fontsize{9.0pt}{\baselineskip}\selectfont
\linespread{0.9}\selectfont
\begin{mybody}
\includegraphics[scale=0.5]{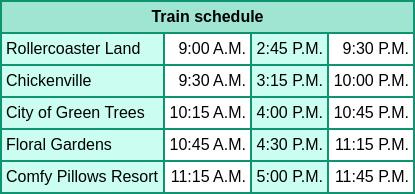}
\\
\textbf{Synthetic Question 1:} What is the travel time from Rollercoaster Land to Chickenville?\\Choices:\\(A) 15 minutes\\(B) 30 minutes\\(C) 45 minutes\\(D) 1 hour \\\\
\textbf{Synthetic Question 2:} What is the travel time from City of Green Trees to Floral Gardens?\\Choices:\\(A) 30 minutes\\(B) 45 minutes\\(C) 1 hour\\(D) 15 minutes\\\\
\textbf{Synthetic Question 3:} What is the most common time interval between consecutive train stops according to the schedule?\\Choices:\\(A) 15 minutes\\(B) 30 minutes\\(C) 45 minutes\\(D) 1 hour
\end{mybody}
\captionof{figure}{\textbf{Qualitative example to show diverse and reasonable synthetic images for a given image.}}
\label{fig:scale_question_example_1}
\end{table*}

\begin{table*}[th!]
\fontsize{9.0pt}{\baselineskip}\selectfont
\linespread{0.9}\selectfont
\begin{mybody}
\includegraphics[scale=0.3]{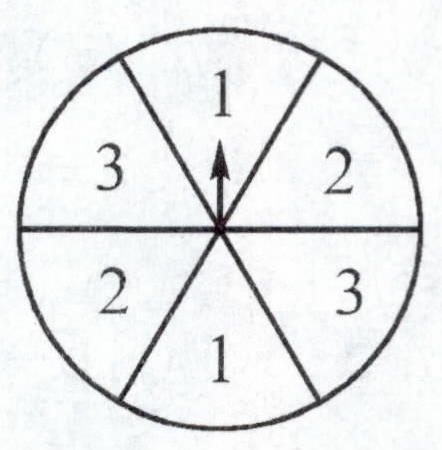}
\\
\textbf{Synthetic Question 1:} What is the probability that the pointer lands on a number greater than 1 when the spinner is spun once? \\\\
\textbf{Synthetic Question 2:} If the spinner is spun twice, what is the probability that the pointer lands on a number greater than 1 on both spins?\\\\
\textbf{Synthetic Question 3:} What is the probability that the pointer lands on a sector labeled with the number 2 when the spinner is spun once?
\end{mybody}
\captionof{figure}{\textbf{Qualitative example to show diverse and reasonable synthetic images for a given image.}}
\label{fig:scale_question_example_2}
\end{table*}

\begin{table*}[th!]
\fontsize{9.0pt}{\baselineskip}\selectfont
\linespread{0.9}\selectfont
\begin{mybody}
\includegraphics[scale=0.5]{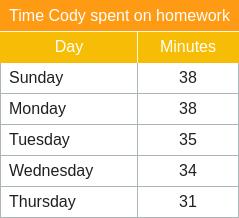}
\\
\textbf{Question:} What is the total time Cody spent on homework from Sunday to Tuesday?\\A) 105 minutes\\B) 111 minutes\\C) 106 minutes\\D) 108 minutes.\\\\
\textbf{Chain-of-Thought:} \\
<caption> The image presents a table illustrating the time Cody spent on homework each day, with data for five days.\\***Table Structure:**\\*The table consists of two columns and six rows.\\*The first row is a header row, with the left column labeled \"Day\" and the right column labeled \"Minutes\".\\*The remaining five rows represent individual days of the week, from Sunday to Thursday.\\***Data:**\\***Sunday:** 38 minutes\\***Monday:** 38 minutes\\***Tuesday:** 35 minutes\\***Wednesday:** 34 minutes\\***Thursday:** 31 minutes\\In summary, the table provides a clear and concise overview of Cody's daily homework time, allowing for easy comparison and analysis of his study habits over the course of a week. </caption>\\
<think>\\
\#\# Step 1: Identify the relevant data\\
To find the total time Cody spent on homework from Sunday to Tuesday, we need to identify the time spent on each of these days. According to the table, Cody spent 38 minutes on Sunday, 38 minutes on Monday, and 35 minutes on Tuesday.\\
\#\# 2: Add up the time spent on each day\\
Now, we add up the time spent on Sunday, Monday, and Tuesday. 
$$
\text{total\_time} = 38 + 38 + 35
$$
\#\# 3: Calculate the total time\\
Perform the addition to find the total time. 
$$
\text{total\_time} = 38 + 38 + 35 = 111 \text{ minutes}
$$
\#\# 4: Evaluate the answer choices\\
Compare the calculated total time to the given answer choices to select the correct one. The choices are A: 105 minutes, B: 111 minutes, C: 106 minutes, D: 108 minutes.\\
\#\# 5: Select the correct answer\\
Based on the calculation, the total time Cody spent on homework from Sunday to Tuesday is 111 minutes, which matches option B.
The steps taken to solve this problem involved directly adding up the time Cody spent on homework from Sunday to Tuesday. The process was straightforward, and the calculation was simple. There were no alternative solutions needed since the question only required a basic arithmetic operation. The reflection on the steps confirms that the approach was correct and efficient for this type of problem.\\
</think>
$$
\boxed{B}
$$
\end{mybody}
\captionof{figure}{\textbf{Qualitative example from \name{} dataset.}}
\label{fig:qual_example_1}
\end{table*}

\begin{table*}[th!]
\fontsize{9.0pt}{\baselineskip}\selectfont
\linespread{0.9}\selectfont
\begin{mybody}
\includegraphics[scale=0.4]{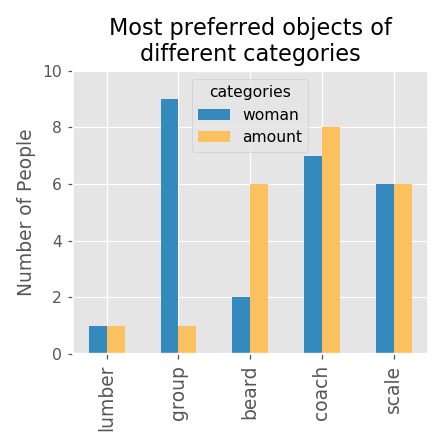}
\\
\textbf{Question:} How many total people preferred the object coach across all the categories?\\\\
\textbf{Chain-of-Thought:} \\
<caption> The image presents a bar graph illustrating the most preferred objects across different categories, with two distinct groups: "woman" and "amount." The graph features five categories on the x-axis: lumber, group, beard, coach, and scale.\\
**Key Observations:**\\
*The y-axis represents the number of people, ranging from 0 to 10.\\
*The graph displays two sets of bars for each category, one in blue (representing "woman") and the other in orange (representing "amount").\\
*Notably, the "group" category has the highest value for "woman," while the "coach" category has the highest value for "amount."\\
**Category-Specific Data:**\\
***Lumber:** 
*Woman: 1
*Amount: 1\\
***Group:** 
*Woman: 9
*Amount: 1\\
***Beard:** 
*Woman: 2
*Amount: 6\\
***Coach:** 
*Woman: 7
*Amount: 8\\
***Scale:** 
*Woman: 6
*Amount: 6\\
**Conclusion:** The graph provides a clear visual representation of the preferred objects across different categories, allowing for easy comparison between the "woman" and "amount" groups. </caption>\\
<think>\\
\#\# Step 1: Identify the relevant data
The image shows a bar graph with different categories on the x-axis and the number of people on the y-axis. The categories are lumber, group, beard, coach, and scale. We are interested in the category "coach".\\
To solve this question, I will first identify the values associated with the "coach" category for both woman and amount. Then, I will add these values to find the total number of people who preferred the object "coach" across all categories.\\
\#\# 2: Extract the values for the coach category
For the "coach" category, the value for woman is 7 and the value for amount is 8.\\
I will now evaluate if these values are correctly identified and if they align with the question asked. The question is about the total number of people who preferred the object "coach", so I need to ensure I'm considering both woman and amount as they represent different aspects of preference.\\
\#\# 3: Calculate the total number of people
To find the total, I add the number of woman who preferred coach (7) and the amount of people who preferred coach (8).\\
$$
\text{total\_people} = 7 + 8 = 15
$$
I will reflect on my approach to ensure it's correct. The approach involves directly adding the values for woman and amount in the coach category. This seems straightforward and aligns with basic principles of data analysis. There are no obvious alternative solutions since the question directly asks for a sum.\\
\#\# 4: Provide the final answer
Given the calculation, the total number of people who preferred the object "coach" across all categories is 15.
</think>
$$
\boxed{15}
$$
\end{mybody}
\captionof{figure}{\textbf{Qualitative example from \name{} dataset.}}
\label{fig:qual_example_2}
\end{table*}

\end{document}